\documentclass[12pt]{article}
\newcommand{\blind}{0}

\usepackage{amsmath}
\usepackage{graphicx}
\usepackage{enumerate}
\usepackage{natbib}
\usepackage{url} 
\usepackage{dsfont}
\usepackage{overpic}
\usepackage{setspace}
\usepackage{enumitem}
\usepackage{pinlabel}    
\usepackage{amssymb}
\usepackage{siunitx}
\usepackage{colortbl}
\usepackage{xcolor}
\usepackage{hyperref}
\usepackage{amsfonts,bm}
\usepackage{amsthm}
\usepackage{microtype}
\usepackage{subfigure}
\usepackage{booktabs} 
\usepackage{multirow} 
\usepackage[toc,page]{appendix}
\usepackage{comment}
\usepackage{caption}
\usepackage{bbm}
\usepackage{algorithm}
\usepackage{algpseudocode}

\DeclareMathOperator*{\argmin}{arg\,min}

\usepackage{hyperref}

\definecolor{ggplotred}{HTML}{F8766D}
\definecolor{ggplotgreen}{HTML}{00BA38}
\definecolor{ggplotblue}{HTML}{619CFF}
\definecolor{plotGreen}{RGB}{0, 128, 0}
\definecolor{plotOrange}{RGB}{255, 165, 0}
\definecolor{plotBlue}{RGB}{0, 0, 255}

\def\vf{{\bm{f}}}
\def\vg{{\bm{g}}}
\def\vh{{\bm{h}}}

\def\vk{{\bm{k}}}

\def\vx{{\bm{x}}}

\def\mH{{\bm{H}}}

\def\mK{{\bm{K}}}

\def\mX{{\bm{X}}}

\def\mBeta{{\bm{\beta}}}
\def\mmu{{\bm{\mu}}}
\def\malpha{{\bm{\alpha}}}

\def\mTheta{{\bm{\Theta}}}
\def\mtheta{{\bm{\theta}}}

\DeclareMathAlphabet{\mathsfit}{\encodingdefault}{\sfdefault}{m}{sl}
\SetMathAlphabet{\mathsfit}{bold}{\encodingdefault}{\sfdefault}{bx}{n}

\def\gL{{\mathcal{L}}}

\def\gR{{\mathcal{R}}}

\def\sR{{\mathbb{R}}}

\begin{document}

\def\spacingset#1{\renewcommand{\baselinestretch}%
{#1}\small\normalsize} \spacingset{1}


\if0\blind
{
  \title{\bf Multi-task Learning for Heterogeneous Multi-source Block-Wise Missing Data}
\author{Yang Sui$^{1}$, Qi Xu$^{2}$, Yang Bai$^{1}$, and Annie Qu$^{3}$\thanks{Corresponding author: \href{mailto:aqu2@uci.edu}{\textcolor{black}{aqu2@uci.edu}}.} 
 \hspace{.2cm} \\
	\textit{$^{1}$School of Statistics and Data Science,}\\
\textit{Shanghai University of Finance and Economics}\\
\textit{$^{2}$Department of Statistics and Data Science, Carnegie Mellon University}\\ 
	\textit{$^{3}$Department of Statistics, University of California, Irvine}} 
  \date{}
  \maketitle
} \fi

\if1\blind
{
  \bigskip
  \bigskip
  \bigskip
  \begin{center}
    {\LARGE\bf Title}
\end{center}
  \medskip
} \fi

\bigskip
\begin{abstract}
Multi-task learning (MTL) has emerged as an imperative machine learning tool to solve multiple learning tasks simultaneously and has been successfully applied to  healthcare, marketing, and biomedical fields. However, in order to borrow information across different tasks effectively, it is essential to utilize both homogeneous and heterogeneous information. Among the extensive literature on MTL, various forms of heterogeneity are presented in MTL problems, such as block-wise, distribution, and posterior heterogeneity. Existing methods, however, struggle to tackle these forms of heterogeneity simultaneously in a unified framework. In this paper, we propose a two-step learning strategy for MTL which addresses the aforementioned heterogeneity. First, we impute the missing blocks using shared representations extracted from homogeneous source across different tasks. Next, we disentangle the mappings between input features and responses into a shared component and a task-specific component, respectively, thereby enabling information borrowing through the shared component. Our numerical experiments and real-data analysis from the ADNI database demonstrate the superior MTL performance of the proposed method compared to other competing methods.
\end{abstract}

\noindent%
\vfill

\newpage
\spacingset{1.75} 
\section{Introduction}\label{sec: intro}
\textbf{Motivation.} 
Many datasets for specific scientific tasks lack sufficient samples to train an accurate machine learning model. In recent decades, multi-task learning (MTL) has become a powerful tool to borrow information across related tasks for improved learning capacity. In addition, data collected for each task might come from multiple sources; for example, clinic notes, medical images, and lab tests are collected for medical diagnosis. The multi-source data brings richer information for each task, potentially enhancing the MTL. However, this also imposes several key challenges. 
First of all, it is common that observed data sources for each task are heterogeneous, so some blocks (certain data sources for certain tasks) could be entirely missing, termed as a block-wise missing structure in the literature. Second, even if the observed data sources are aligned across tasks, the distribution of the same data source could be heterogeneous, referred to as distribution heterogeneity. Furthermore, the associations between features and responses could vary due to distinct scientific goals or other factors, which we refer to as posterior heterogeneity. In the following, we provide concrete motivating examples to illustrate these challenges in different problems.

\textbf{Example 1: Medical multi-source datasets.} 
Multi-source data are widely observed in medical applications and offer more comprehensive information than single-source data. For example, the Alzheimer’s Disease Neuroimaging Initiative (ADNI) dataset includes medical imaging, biosamples, gene expression, and demographic information \citep{mueller2005alzheimer, mueller2005ways}. However, entire blocks of data are often missing when certain sources become unnecessary or infeasible to collect due to known factors or patient conditions \citep{madden2016missing}.

\textbf{Example 2: Single-cell multi-omics datasets.} Data from different experimental batches often exhibit distribution heterogeneity across various omics measurements. For instance, transcriptome data collected from different batches can display varying patterns due to differences in experimental conditions or technical variability \citep{cao2022unified, cao2022manifold}. In multi-omics datasets, sequencing data distributions also differ across various cancer types \citep{subramanian2020multi}.

\textbf{Example 3: Combining randomized controlled trials (RCTs) and observational data.} Combining RCTs and observational data has become effective for deriving causal effects due to the high costs and limited participant numbers in RCTs \citep{colnet2024causal}. However, RCTs and observational data often exhibit posterior heterogeneity \citep{li2024combining}; for instance, causal effects in RCTs may differ from associations in observational data due to the controlled conditions of RCTs \citep{imbens2015causal}.

\textbf{Challenges.} The challenge in MTL is to incorporate various forms of heterogeneity, each introducing a unique challenge. Block-wise heterogeneity complicates the integration of data as missing patterns vary across tasks, making it difficult to leverage shared information efficiently. For example, in the ADNI dataset, imaging features are present in all datasets, but genetic information is available only in specific subsets \citep{xue2021integrating}. In addition, distribution heterogeneity can also lead to biased or misleading scientific conclusions if not addressed properly. For instance, in multi-omics datasets, sequencing data vary significantly across different cancer types \citep{subramanian2020multi}. Lastly, posterior heterogeneity affects the accuracy of predictions. For example, the relationships identified in RCTs often do not align with those observational data collected in real-life settings \citep{kent2018personalized,kent2020predictive}. While each type of heterogeneity imposes its own challenge, addressing all three challenges simultaneously under a unified framework presents significant obstacles, and to our best knowledge, current MTL methods are not equipped to handle these intricate dilemmas.

\textbf{Contributions.} In this work, we propose a unified MTL framework to address three types of heterogeneity in MTL. There are three key contributions: First, we propose a novel block-wise missing imputation method which effectively handles distribution heterogeneity by learning both shared and task-specific representations, uncovering complex structures between sources, and enabling better generalization during imputation. Second, we disentangle the associations between all input features and responses into shared and task-specific components, allowing for the effective integration of information while adapting to differences across tasks. Third, we propose an MTL architecture consisting of two parts to construct these associations. The first part builds heterogeneous feature spaces, while the second part learns responses, jointly addressing both distribution and posterior heterogeneity. 
We validate the proposed framework on synthetic and real-world datasets, demonstrating its superior performance in handling block-wise missing data and various levels of heterogeneity.
\section{Related Work}\label{sec: related work}
\textbf{Multi-source data integration.} Several related works on multi-source data collected for the same set of samples fall within the Joint and Individual Variation Explained (JIVE) framework. These methods are classified as unsupervised or supervised JIVE, depending on the presence of responses. Unsupervised JIVE and its variants learn joint, individual, and partially shared structures from multiple data matrices through low-rank approximations \citep{lock2013joint, feng2018angle, gaynanova2019structural, choi2022integrative, yi2023hierarchical, james2024learning}. Supervised JIVE, on the other hand, focuses on regression for multi-source data \citep{gao2021covariate, palzer2022sjive, zhang2022joint, wang2024multiple}. Similarly, factor models have been applied to multi-source data in a supervised setting \citep{shu2020d, li2022integrative, anceschi2024bayesian}. While these methods can effectively address distribution heterogeneity across different sources in linear settings, they are limited in scope, as they capture only simple data structures within a single task.

\textbf{Multi-source block-wise missing data integration.} Recently, several methods have been developed to address block-wise missing data. These methods can be divided into two categories based on whether imputation is involved. Imputation-based methods assume consistent correlations between different sources across datasets, allowing for the imputation of missing blocks \citep{gao2017high, le2021sa,xue2021integrating, xue2021statistical, zhou2021multi, ouyang2024imputation}. For example, \citet{xue2021integrating} and \citet{xue2021statistical} construct estimating equations using all available information and integrate informative estimating functions to achieve efficient estimators. On the other hand, non-imputation-based methods focus on learning the covariance matrices among predictors and between the response and predictors from the observed blocks \citep{yuan2012multi, xiang2014bi, yu2020optimal, li2024adaptive}. While these methods perform well in the absence of distribution shift and posterior shift, effectively utilizing all block-wise missing data, they struggle to handle distribution or posterior heterogeneity.

\textbf{Multi-task learning (MTL).} There is a growing literature on learning multiple tasks simultaneously with a shared model; see \citet{zhang2018overview, crawshaw2020multi, zhang2021survey} for reviews. Here, we primarily focus on MTL with deep neural networks, as these networks can capture more complex relationships. These methods can be broadly classified into four categories: The first category is balancing individual loss functions for different tasks, which is a common approach to ease multi-task optimization \citep{du2018adapting, gong2019comparison, hang2023efficient, wu2024adaptive}. The second category involves regularization, especially in the form of hard parameter sharing \citep{subramanian2018learning, liu2019multi, maziarka2022relationship} and soft parameter sharing \citep{ullrich2017soft, lee2018joint, han2024parameter}. The third category addresses the challenge of negative transfer, where explicit gradient modulation is used to alleviate conflicts in learning dynamics between tasks \citep{lopez2017gradient, chaudhry2018efficient, maninis2019attentive, abdollahzadeh2021revisit, hu2022improving, wang2024comprehensive}. The fourth category uses knowledge distillation to transfer knowledge from single-task networks to a multi-task student network \citep{rusu2015policy, teh2017distral, clark2019bam, d2024sharing}. Although MTL can integrate data from multiple tasks, it is limited in addressing different types of heterogeneity and is constrained by the assumption of a fully observed setting.

Most related work focuses on addressing a single challenge, such as posterior heterogeneity or the missing data problem, but typically fails to address all challenges simultaneously. In contrast, our proposed method extends these approaches by tackling both distribution and posterior heterogeneity in a block-wise missing setting. This enables a more comprehensive integration of data across tasks, resulting in improved performance in MTL.

\section{Two-step MTL Framework}
\textbf{Notation.} We introduce the notations used in this paper. Vectors and matrices are denoted by $\vx$ and $\mX$, respectively. The $\ell_1$ and $\ell_2$ norms of a vector $\vx$ are $\|\vx\|_1$ and $\|\vx\|_2$, and the Frobenius norm of a matrix $\mX$ is $\|\mX\|_F$. The symbol $|$ represents concatenation. For example, $[\vx_1 | \vx_2]$ denotes concatenating a $p_1 \times 1$ vector $\vx_1$ and a $p_2 \times 1$ vector $\vx_2$ into a $(p_1 + p_2) \times 1$ vector. Similarly, $[\mX_1 | \mX_2]$ denotes concatenating an $n \times p_1$ matrix $\mX_1$ and an $n \times p_2$ matrix $\mX_2$ into an $n \times (p_1 + p_2)$ matrix. We define $[r] = \{1, 2, \ldots, r\}$ as the set of integers from $1$ to $r$.

\begin{figure}[t]
\begin{center}
        \begin{overpic}[width=0.5\linewidth]{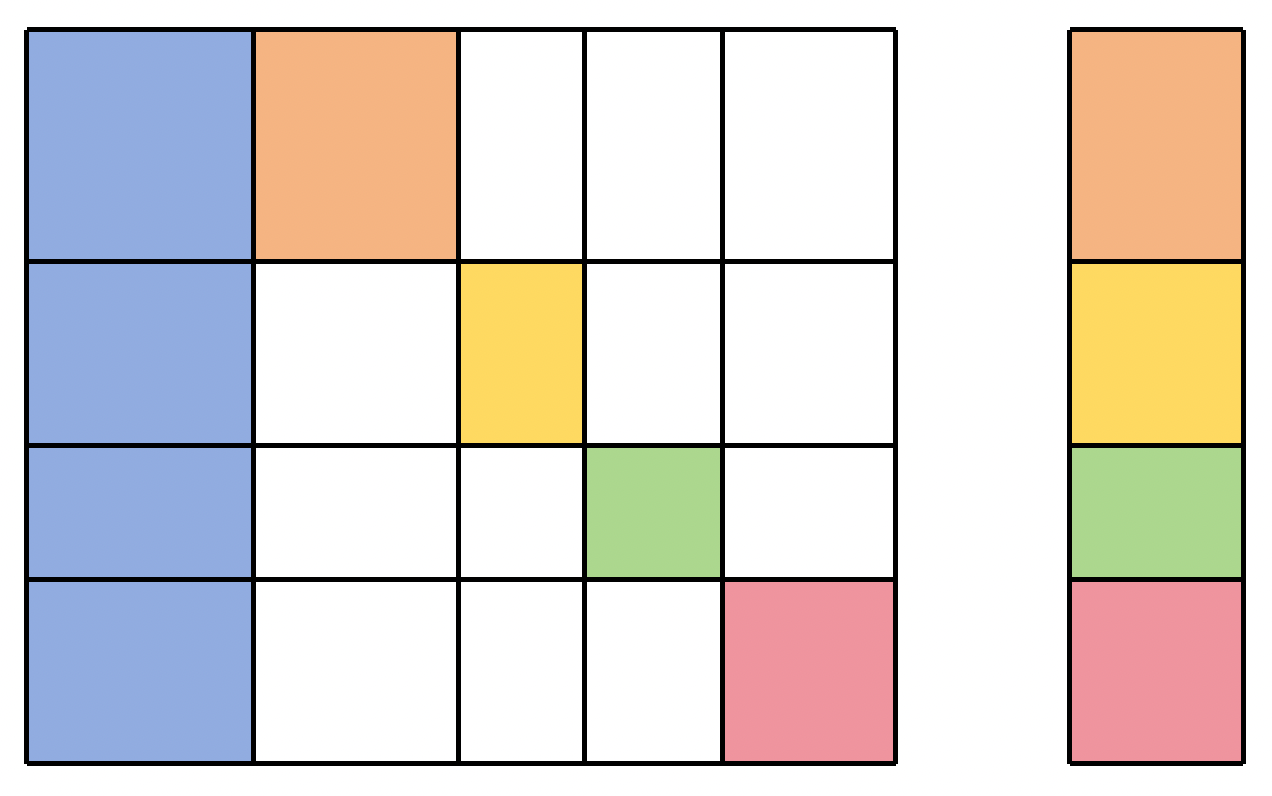}
        \put(3,62){\textit{\color{blue}Anchor}}
        \put(8.5,49){$\vx^1_0$}
        \put(8.5,33){$\vx^2_0$}
        \put(8.5,21){$\vx^3_0$}
        \put(8.5,9){$\vx^4_0$}
        \put(25,62){\textit{\color{red}Specific Sources}}
        \put(25,49){$\vx^1_1$}
        \put(38,33){$\vx^2_2$}
        \put(48,21){$\vx^3_3$}
        \put(61,9){$\vx^4_4$}

        \put(77,62){\textit{Response}}
        \put(89,49){$y^1$}
        \put(89,33){$y^2$}
        \put(89,21){$y^3$}
        \put(89,9){$y^4$}

    \end{overpic}
    \caption{Block-wise missing pattern for \num{4} tasks and \num{5} sources, including an anchoring source and task-specific sources.}
    \label{fig:block_missing_pattern}
\end{center}
\end{figure}

\textbf{Problem Description.} Suppose we have data from $T$ tasks, with features collected from $T+1$ sources. For all tasks, we assume that a common source, called the anchoring source, is observed. Additionally, each task has its own task-specific source, denoted as $\vx^t_s$ for the $s$-th source in the $t$-th task. Specifically, $\vx^t_0$ represents the anchoring source observed in the $t$-th task, and $\vx^t_t$ denotes the task-specific source for the $t$-th task, while $\{\vx^t_s\}_{s \neq 0, t}$ are missing. For the $t$-th task, we observe $n_t$ samples $\{[\vx^t_{0,i} | \vx^t_{t,i}], y^t_i\}_{i=1}^{n_t}$. This block-wise missing pattern is common in real-world applications. For example, in biomedical data, some measurements (data sources) are widely observed for all subjects, while some measurements are only collected to a subgroup of subjects due to various reasons. Concretely, in the ADNI data that we analyzed in Section \ref{sec: adni}, MRI is crucial to monitor the cognitive impairment development of Alzheimer’s patients, so it is measured for all subjects, while gene expression and PET images are less crucial and are only observed for two subgroups separately.
Another example is the split questionnaire design, which aims to reduce respondent fatigue and improve response rates by assigning different subsets of the questionnaire to different sampled respondents \citep{lin2023optimal}. In Figure~\ref{fig:block_missing_pattern}, we provide an example of a block-wise missing pattern for \num{4} tasks and \num{5} sources, where the blue source $\vx^t_0$ for $t \in [4]$ represents the anchoring source observed by all four tasks, and each task also has a uniquely observed specific source $\vx^t_t$ for $t \in [4]$. Our goal is to perform MTL on these tasks with block-wise missing data.

Figure~\ref{fig:block_missing_pattern} illustrates one of the challenges in MTL. Each task has different missing blocks; for example, in the first task, sources \num{2}, \num{3}, and \num{4} are missing, while in the second task, sources \num{1}, \num{3}, and \num{4} are missing. Furthermore, both distribution and posterior heterogeneity across tasks complicate the application of standard imputation methods \citep{nair2019covariate, he2024transfusion, he2024adatrans} and MTL methods \citep{kouw2018introduction, lee2024enhancing, maity2024linear}.

\subsection{Heterogeneous Block-wise Imputation}\label{sec: imputation}
In this section, we propose the first step, Heterogeneous Block-wise Imputation (HBI) for imputing the missing blocks while leveraging distribution heterogeneity across tasks. HBI extracts disentangled hidden representations from the anchoring source $\vx_{0}$, including a shared representation across tasks and a task-specific representation for each task. The shared representation is then used to impute the missing blocks, improving generalization across tasks.

\begin{figure}[t]
\begin{center}
        \begin{overpic}[width=0.5\linewidth]{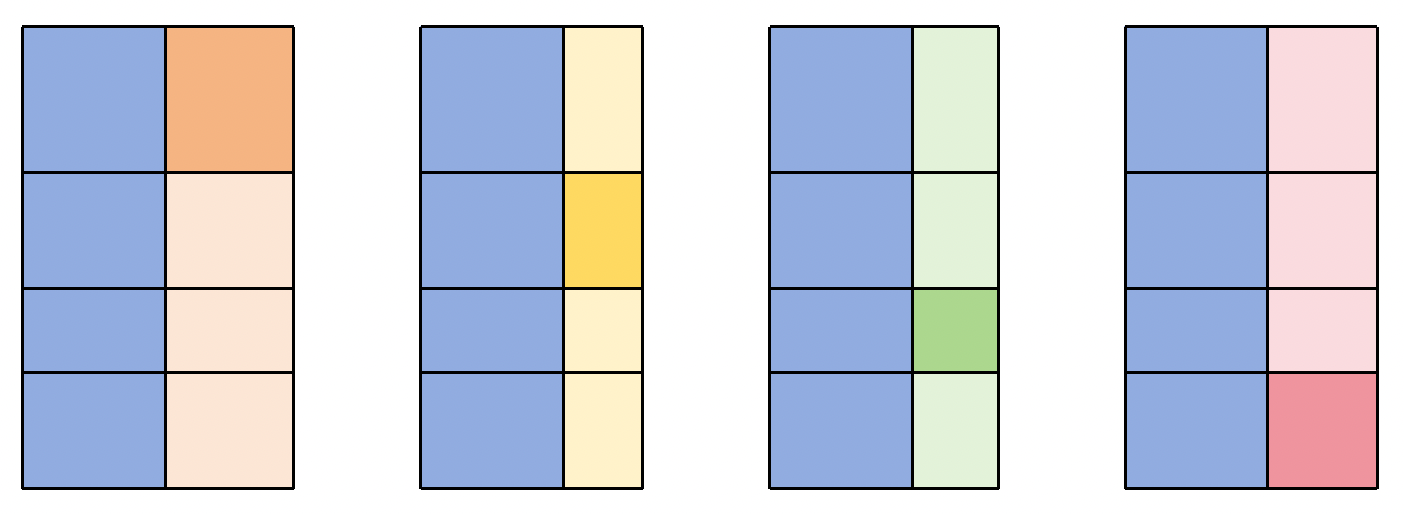}
        \put(4,28){$\vx^1_0$}
        \put(4,18.5){$\vx^2_0$}
        \put(6,11.5){$\cdot$}
        \put(4,4.5){$\vx^4_0$}
        \put(14,28){$\vx^1_1$}

        \put(32,28){$\vx^1_0$}
        \put(32,18.5){$\vx^2_0$}
        \put(34,11.5){$\cdot$}
        \put(32,4.5){$\vx^4_0$}
        \put(40,18.5){$\vx^2_2$}

        \put(57,28){$\vx^1_0$}
        \put(57,18.5){$\vx^2_0$}
        \put(59,11.5){$\cdot$}
        \put(57,4.5){$\vx^4_0$}
        \put(67,11.5){$\cdot$}

        \put(82.5,28){$\vx^1_0$}
        \put(82.5,18.5){$\vx^2_0$}
        \put(84.5,11.5){$\cdot$}
        \put(82.5,4.5){$\vx^4_0$}
        \put(91.5,4.5){$\vx^4_4$}
    \end{overpic}
    \caption{Illustration of parallel imputation for task-specific sources.}
    \label{fig:parallel}
\end{center}
\end{figure}
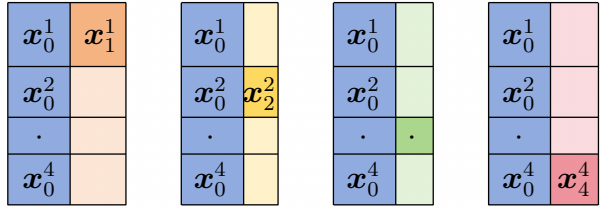

For $T$ tasks and $T+1$ sources, we impute the task-specific sources in a parallel fashion. For each task-specific source $s\neq0$, we utilize the anchoring source across all tasks and $\vx^{s}_{s}$ to impute the unobserved blocks $\{\vx^{t}_{s}\}_{t \neq s}$. In particular, for the $t$-th source, only the $t$-th task has observed values for the features $\vx^t_t$. The imputation aims to use the observed $\vx^t_{0}$ and $\vx^t_t$ along with $\vx^{-t}_{0} = \{\vx^r_0\}_{r \neq t}$ to estimate the missing features in the $t$-th source for the other $T-1$ tasks, where $\vx^{-t}_{t} = \{\vx^{r}_{t}\}_{r \neq t}$ are unobserved. For example, in Figure~\ref{fig:parallel}, we use information from $\vx^1_0$, $\vx^1_1$, and $\vx^{-1}_0 = \{\vx^2_0, \vx^3_0, \vx^4_0\}$ to impute the missing blocks $\vx^{-1}_1 = \{\vx^2_1, \vx^3_1, \vx^4_1\}$ for the task \num{1}-specific source.

\begin{figure}[t]
\begin{center}
        \begin{overpic}[width=0.5\linewidth]{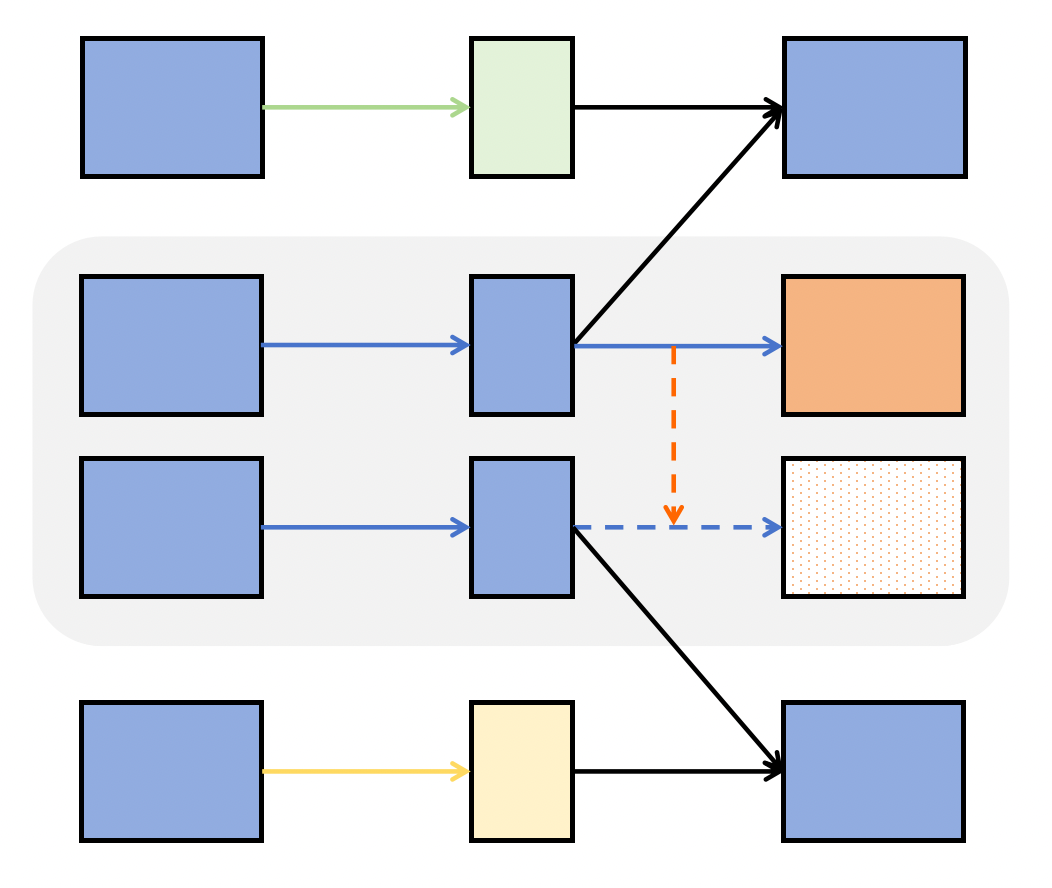}
        \put(13,73){$\vx^t_0$}
        \put(13,50){$\vx^t_0$}
        \put(13,33){$\vx^{-t}_0$}
        \put(13,10){$\vx^{-t}_0$}

        \put(33,77){$E_p^t$}
        \put(33,54){$E_c$}
        \put(33,37){$E_c$}
        \put(33,14){$E_p^{-t}$}

        \put(47,73){$\vg^t$}
        \put(47,50){$\vf^t$}
        \put(45,33){$\vf^{-t}$}
        \put(45.5,10){$\vg^{-t}$}

        \put(61,77){$D$}
        \put(61,54){$G$}
        \put(61,14){$D$}

        \put(80,73){$\widehat\vx^t_0$}
        \put(80,50){$\widehat\vx^t_t$}
        \put(80,10){$\widehat\vx^{-t}_0$}

    \end{overpic}
    \caption{Illustration of HBI for the task $t$-specific source $\vx_t$. A common encoder $E_c(\cdot)$ learns to capture representation components that are shared among tasks. Task-specific encoders $E_p(\cdot)$ (one for the $t$-th task, and one for the other $T-1$ tasks) learn to capture task-specific components of the representations. A decoder learns to reconstruct the anchoring source $\vx_0$ by using both shared and task-specific representations. The shared part of the relationship between the anchoring source $\vx_0$ and the task $t$-specific source $\vx_t$ can be borrowed through $E_c(\cdot)$ and $G(\cdot)$ for imputation. See the text for more information.}
    \label{fig:imputation}
\end{center}
\end{figure}
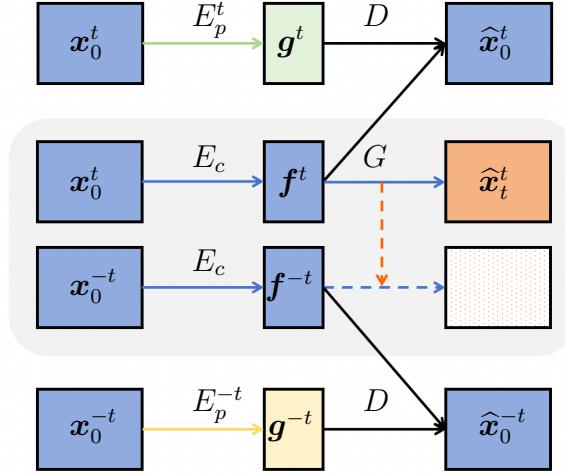

This is accomplished by learning a model that exploits both the shared and task-specific information of the data, allowing for accurate prediction of missing values based on the available observed data. To fully integrate multi-source information, we leverage an encoder-decoder framework, which is well-suited for capturing non-linear relationships in data. Let $E_c(\cdot)$ be a common encoder that maps $\{\vx^t_0, \vx^{-t}_0\}$ to shared representations $\vf^t_c = E_c(\vx^t_0)$ and $\vf^{-t}_c = E_c(\vx^{-t}_0)$ across all $T$ tasks. Let $E_p^{t}(\cdot)$ and $E_p^{-t}(\cdot)$ be task-specific encoders that map $\vx_0^t$ and $\vx^{-t}_0$ to task-specific representations $\vg^t = E_p^{t}(\vx^t_0)$ and $\vg^{-t} = E_p^{-t}(\vx^{-t}_0)$. Then, $D(\vf, \vg)$ serves as a decoder that reconstructs the anchoring source $\vx_0$ from $\vf$ and $\vg$. Finally, $G(\vf)$ is a predictor that maps the shared representation $\vf$ to the task $t$-specific source $\vx_{t}$. The resulting heterogeneous block-wise imputation model is illustrated in Figure~\ref{fig:imputation}.

In Figure~\ref{fig:imputation}, we assume that the relationship between the anchoring source $\vx_{0}$ and the task $t$-specific source $\vx_{t}$ can be borrowed through the shared representations $\vf$, the common encoder $E_c(\cdot)$, and the decoder $G(\cdot)$. This allows us to utilize the shared information (reflected in $\vf^t$ and $\vf^{-t}$) for imputation, while also accounting for the heterogeneity between $\vx^t_0$ and $\vx^{-t}_{0}$ (reflected in $\vg^t$ and $\vg^{-t}$). 
Existing imputation methods often learn the relationship between $\vx_0$ and $\vx_t$ within the $t$-th task and apply the relationship to other tasks, overlooking distribution heterogeneity \citep{xue2021statistical, zhou2021multi}. Moreover, common imputation methods rely on parametric models which fail to capture complex relationships in missing data \citep{xue2021integrating, li2023weighted}. However, our HBI method effectively overcomes these obstacles. Notably, HBI’s extraction of the common components in the relationships between sources across different tasks shares similarities with domain adaptation \citep{mansour2008domain, bousmalis2016domain, tzeng2017adversarial, farahani2021brief} but focuses on completing block-wise missing data. This architecture effectively models complex data structures and interactions, providing a robust tool for understanding intricate patterns. The resulting optimization can be formulated as:
\begin{align}\label{eqn:imputation loss}
\begin{split}
    \{\widehat E_c(\cdot),\widehat E^{t}_p(\cdot), \widehat E^{-t}_p(\cdot),\widehat D(\cdot),\widehat G(\cdot)\}=\argmin\{\gL_{\text{pre}}+\gL_{\text{recon}}\},
\end{split}
\end{align}
In \eqref{eqn:imputation loss}, the prediction loss $\gL_{\text{pre}}$ trains the model to predict $\vx^t_t$, the target of interest, which is applied only to the $t$-th task. We use the following loss function:
$$
\gL_{\text{pre}}=\sum_{i=1}^{n_t}l(\vx^t_{t,i}, G(E_c(\vx^t_{0,i}))),
$$
where $\vx_i$ denotes the observed sample, and $l(\cdot, \cdot)$ can be the mean squared error for continuous outcomes or cross-entropy for binary outcomes (this applies similarly to the following symbols). For the reconstruction loss in \eqref{eqn:imputation loss},
\begin{align*}
 \gL_{\text{recon}}=\sum_{i=1}^{n_t}l(\vx^t_{0,i}, D(E_c(\vx^t_{0,i}),E^{t}_p(\vx^t_{0,i})))+\sum_{i=1}^{n_{-t}}l(\vx^{-r}_{0,i}, D(E_c(\vx^{-t}_{0,i}),E^{-t}_p(\vx^{-t}_{0,i}))),   
\end{align*}
where $n_{-t} = \sum_{r \neq t} n_r$. 
Then, we can train \eqref{eqn:imputation loss} to obtain the estimators $\widehat{E}_c(\cdot)$ and $\widehat{G}(\cdot)$. Consequently, we compute  $\widehat{\vx}{}^{-t}_{t} = \widehat{G}(\widehat{E}_c(\vx^{-t}_{0}))$. Note that \eqref{eqn:imputation loss} is constructed based on task $t$-specific source imputation. Similarly, we can construct imputations for the other $T-1$ sources. When performing imputation for different sources using HBI, the learned hidden representations and corresponding generative functions adapt dynamically. This adaptation is crucial as it allows the model to accommodate the unique information of each source. The complete algorithm for parallel heterogeneous imputation is provided in the Supplementary Material \ref{app: algorithm}.

Our proposed HBI method ensures that the imputation model leverages common information across tasks while incorporating the heterogeneity of each task. By decomposing the latent space into shared and task-specific components, we gain a nuanced understanding of how input features from different sources interact, thereby enhancing imputation accuracy.

\subsection{Heterogeneous Multi-task Learning}\label{sec: integration}
In this section, we propose our MTL framework to accommodate distribution and posterior heterogeneity given the imputed blocks from HBI. Similar to the disentangled representations for features, we also model the association between features and responses as two components: a shared function mapping and a task-specific function mapping. Specifically, for the $t$-th task, we assume that the relationship between the response $y^t$ and the features $[\vx^t_0|\vx^t_1|\cdots|\vx^t_T]$ is given by:
\begin{align}\label{eqn:pathway to y}
y^t = \psi_c([\vx^t_0|\vx^t_1|\cdots|\vx^t_T]) + \psi^{t}_{p}([\vx^t_0|\vx^t_1|\cdots|\vx^t_T]),
\end{align}
where $[\vx^t_0|\vx^t_1|\cdots|\vx^t_T]$ influence $y^t$ through a shared mapping $\psi_c(\cdot)$ and a task-specific mapping $\psi_p^t(\cdot)$. Equation \eqref{eqn:pathway to y} extends traditional meta-analysis, which often assumes a linear relationship in the $t$-th task as $y^t = [\vx^t_0|\vx^t_1|\cdots|\vx^t_T]^\top \mBeta^{t} + \varepsilon$, where $\mBeta^{t}$ includes a common component $\mmu$ shared across all $T$ tasks and a unique component $\malpha^{t}$ for each task \citep{chen2021identifying, cai2022individual}. 
Traditional meta-analysis is incapable of accommodating non-linear relationships or varying effects. In contrast, we propose a flexible framework which accommodates non-linearities and integrates task-specific information.

To construct the shared mapping $\psi_c(\cdot)$ and the task-specific mappings $\{\psi_p^t\}_{t=1}^T$ jointly, we consider an MTL architecture comprising two parts. The first part builds heterogeneous feature spaces, while the second part learns responses for all $T$ tasks. Specifically, for the $t$-th task, following HBI in Section \ref{sec: imputation}, we obtain samples with reconstructed features $\{(\vx^t_{0,i}, \ldots, \widehat{\vx}^t_{t-1,i}, \vx^t_{t,i}, \widehat{\vx}^t_{t+1,i}, \ldots, \widehat{\vx}^t_{T,i}), y^t_i\}_{i=1}^{n_t}$. 
These features can then be integrated to capture both shared and task-specific representations, enabling the utilization of the combined data while addressing task-specific heterogeneity.
During HBI, the components $\{\widehat{\vx}^t_{s}\}_{s \neq 0, t}$ are primarily predicted using the anchoring source $\vx^t_0$, indicating that $\vx^t_0$ serves as a common basis. To prevent redundancy, we extract shared representations solely from the anchoring source $\vx_0^t$. Specifically, we define
\begin{align}\label{eqn:common features}
   \vh^{t} = \phi_c(\vx^t_0), 
\end{align} 
where $\phi_c(\cdot)$ is a shared encoder that learns hidden information from the anchoring source for all tasks. Meanwhile, task heterogeneity is captured by extracting representations from all features, creating a framework in which shared representations provide a 
foundation, while task-specific details can still be preserved. For the $t$-th task, we define:
\begin{align}\label{eqn:private features}
\vk^{t} = \phi^{t}_p([\vx^t_0|\cdots|\widehat \vx^t_{t-1}|\vx^t_t|\widehat \vx^t_{t+1}|\cdots|\widehat \vx^t_T]),
\end{align}
where $\phi^{t}_p$ is a task-specific encoder which maps the unique information within the $t$-th task. In \eqref{eqn:common features} and \eqref{eqn:private features}, the heterogeneous feature spaces are fully captured using all data information. 
The task-specific representations $\vk$ capture complex interactions between different sources unique to each task, aided by HBI in Section \ref{sec: imputation}.
In practice, such interactions are crucial. For example, in the ADNI dataset, there are intricate relationships between images and gene expression. Equation \eqref{eqn:private features} accounts for this heterogeneous information. However, previous work \citep{moon2017completely, bica2022transfer} often oversimplifies these interactions by focusing only on task-specific sources, neglecting a wealth of shared information from other tasks.

For the second part, we consider a network architecture for learning responses in all $T$ tasks, consisting of $L$ layers with both shared and task-specific subspaces \citep{curth2021inductive, bica2022transfer}. For simplicity, in the $t$-th task, let $\bar{\vk}^t_{l}$ and $\bar{\vh}^t_{l}$ represent the inputs, and $\vk^t_{l}$ and $\vh^t_{l}$ the outputs of the $l$-th layer. For $l=1$, set $\bar{\vk}^t_{1} = [\vh^t|\vk^t]$ and $\bar{\vh}^t_{1} = \bar{\vk}^t_{1}$. For $l > 1$, the inputs to the $(l+1)$-th layer are given by $\bar{\vk}^t_{l+1} = [\vh^t_{l}|\vk^t_{l}]$ and $\bar{\vh}^t_{l} = [\vh^t_{l}]$. 
Let $g^t(\cdot)$ be the association function in the $t$-th task, defined as $g^t([\vh^t_{L}|\vk^t_{L}]) = \psi_c([\vx^t_0|\vx^t_1|\cdots|\vx^t_T]) + \psi^{t}_{p}([\vx^t_0|\vx^t_1|\cdots|\vx^t_T])$, where $g^t(\cdot)$ is a linear function for continuous outcomes and a sigmoid function for binary ones.

\begin{figure}[t]
\begin{center}
        \begin{overpic}[width=0.5\linewidth]{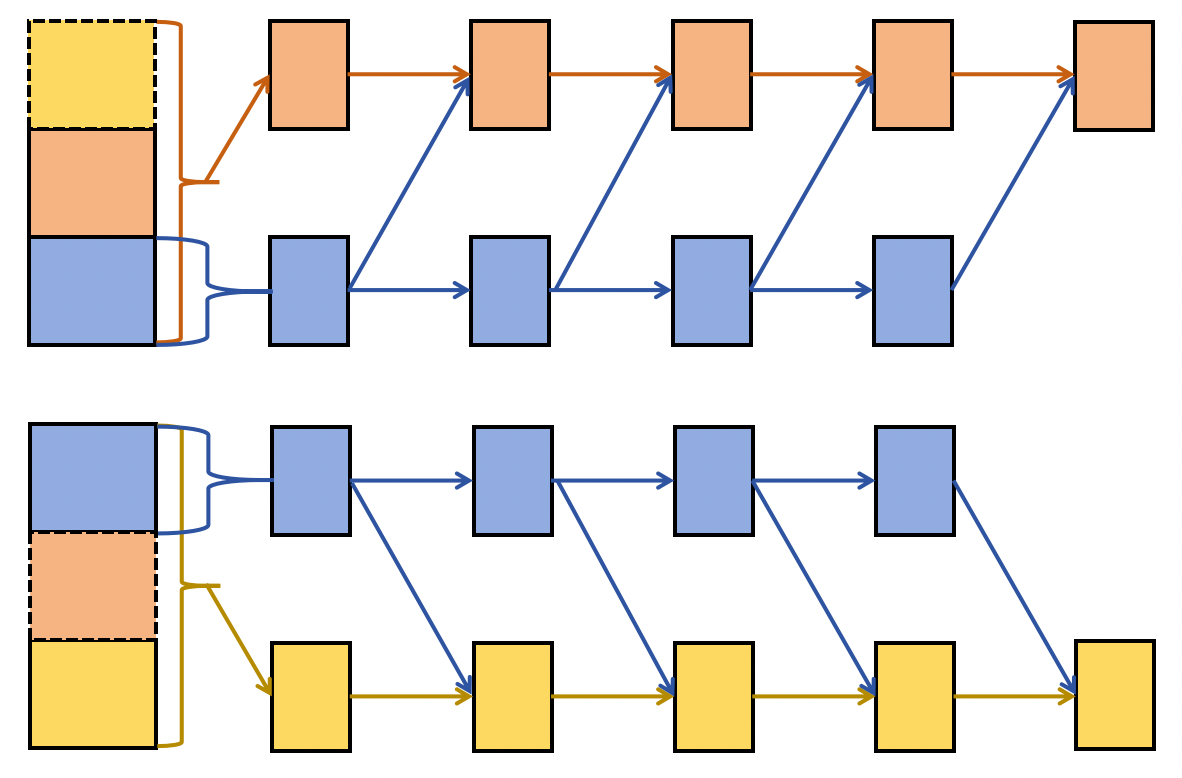}
        \put(1,66){\textit{task 1}}
        \put(16,60){$\phi^1_p$}
        \put(5,58){$\widehat\vx^1_2$}
        \put(5,49){$\vx^1_1$}
        \put(5,40){$\vx^1_0$}
        \put(16,32){$\phi_c$}

        \put(5,24){$\vx^2_0$}
        \put(5,15){$\widehat\vx^2_1$}
        \put(5,6){$\vx^2_2$}
        \put(16,4){$\phi^2_p$}
        \put(1,-3){\textit{task 2}}

        \put(23.5,58){$\vk^1$}
        \put(23.5,40){$\vh^1$}

        \put(23.5,24){$\vh^2$}
        \put(23.5,6){$\vk^2$}

        \put(57.7,58){$\vk^1_l$}
        \put(57.7,40){$\vh^1_l$}
        \put(57.7,24){$\vh^2_l$}
        \put(57.7,6){$\vk^2_l$}

        \put(92,58){$y^1$}
        \put(92,6){$y^2$}
    \end{overpic}
    \caption{Illustration of the construction of shared mapping and task-specific mappings for two tasks.}
    \label{fig:mtl}
\end{center}
\end{figure}
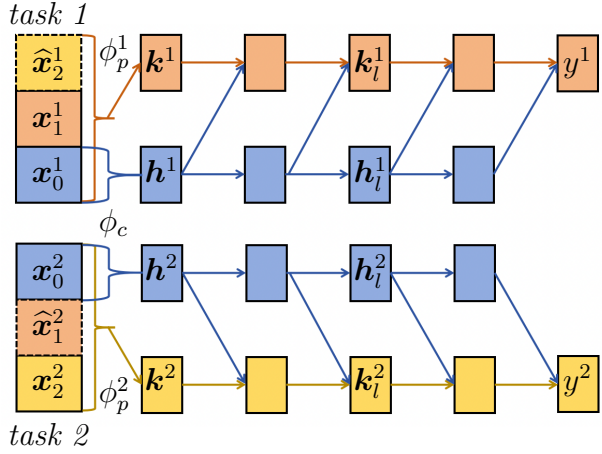

Figure~\ref{fig:mtl} illustrates the construction of the shared mapping $\psi_c(\cdot)$ and the task-specific mappings $\psi_p^1(\cdot)$ and $\psi_p^2(\cdot)$ for two tasks. For task \num{1}, the input features consist of $[\vx^1_0|\vx^1_1|\widehat{\vx}^1_2]$, where $\widehat{\vx}^1_2$ represents the imputed source. In the first part, we build the heterogeneous feature space by extracting the shared representation $\vh^1 = \phi_c(\vx^1_0)$ and the task-specific representation $\vk^1 = \phi_p^1([\vx^1_0|\vx^1_1|\widehat{\vx}^1_2])$. Similarly, for task \num{2}, we extract $\vh^2 = \phi_c(\vx^2_0)$ and $\vk^2 = \phi_p^2([\vx^2_0|\widehat{\vx}^2_1|\vx^2_2])$. 
Next, we utilize the pairs $\{\vh^1, \vk^1\}$ and $\{\vh^2, \vk^2\}$ to model the responses $y^1$ and $y^2$, respectively. In Figure~\ref{fig:mtl}, the blue mapping illustrates the shared mapping $\psi_c(\cdot)$, while the orange and yellow mappings represent the task-specific mappings $\psi_p^1(\cdot)$ and $\psi_p^2(\cdot)$ for tasks \num{1} and \num{2}, respectively. 

The above construction allows us to define the following integrated loss across all $T$ tasks:
\begin{align}\label{eqn:loss integ}
\gL_{\text{integ}} = \sum_{t=1}^T \sum_{i=1}^{n_t} l(y^t_i, g^t([\vh^t_{L,i}|\vk^t_{L,i}])).
\end{align}
Similar to \citet{bousmalis2016domain}, we also incorporate an orthogonality regularizer, defined as:
\begin{align}\label{eqn:loss orth}
\gR_{\text{orth}} = \sum_{t=1}^T \|(\mH^t)^{\top} \mK^t\|^2_F,
\end{align}
where $\mH^t$ and $\mK^t$ are matrices whose rows are
$\vh^t$ and $\vk^t$, respectively. Furthermore, in \eqref{eqn:private features}, the input is $[\vx^t_0 \,|\, \cdots \,|\, \widehat{\vx}^t_{t-1} \,|\, \vx^t_t \,|\, \widehat{\vx}^t_{t+1} \,|\, \cdots \,|\, \widehat{\vx}^t_{T}]$, where $\vx^t_0$ and $\vx^t_t$ are the observed data, and $\{\widehat{\vx}^t_{s}\}_{s \neq 0, t}$ are obtained through imputation. Since imputation can introduce errors, we downweight the imputed data $\{\widehat{\vx}^t_{s}\}_{s \neq 0, t}$ compared to observed data for learning $\vk^t$ by applying a regularizer to the parameters of the first layer of the encoder $\phi^t_p(\cdot)$, defined as:
\begin{align}\label{eqn:loss rbt}
\gR_{\text{imp}} = \sum_{t=1}^T \sum_{s\neq 0, t}\|\mTheta^t_{s,p,1}\|^2_F,
\end{align}
where $\mTheta^t_{s,p,1}$ are the parameters of the first layer of $\phi^t_p(\cdot)$ corresponding to $\{\widehat{\vx}^t_s\}_{s \neq 0, t}$. This regularizer downweights potentially less accurate imputed features by penalizing the magnitude of the encoder parameters, fostering a model more robust to imputation errors.
To further reduce redundancy between the shared and task-specific layers, we introduce an orthogonal regularizer \citep{ruder2019latent}. Let $d^t_{c, l-1}$ and $d^t_{p, l-1}$ be the dimensions of $\vh^t_{l-1}$ and $\vk^t_{l-1}$, the outputs of the $(l-1)$-th layer. Denote the weights in the $l$-th layer as $\mTheta^t_{c, l} \in \sR^{d^t_{c, l-1} \times d^t_{c, l}}$ and $\mTheta^t_{p, l} \in \sR^{(d^t_{c, l-1} + d^t_{p, l-1}) \times d^t_{p, l}}$. We apply the following regularizer:
\begin{align}\label{eqn:loss dr}
\gR_{\text{dr}} = \sum_{t=1}^T \sum_{l=1}^L \| (\mTheta^t_{c,l})^\top\mTheta^t_{p,l,1:d^t_{c,l-1}}\|_F^2.
\end{align}
By combining the losses from \eqref{eqn:loss integ}, \eqref{eqn:loss orth}, \eqref{eqn:loss rbt}, and \eqref{eqn:loss dr}, we train the set of all parameters $\mTheta$ using the following integrated loss function:
$$
\{\widehat{\psi}_c, \{\widehat{\psi}^t_p\}_{1 \le t \le T}\} = \argmin_{\mTheta} \{\gL_{\text{integ}} + \gamma \gR_{\text{orth}} + \delta \gR_{\text{imp}} + \kappa \gR_{\text{dr}}\},
$$
where $\gamma$, $\delta$, and $\kappa$ are weights controlling the balance among different terms. The more detailed algorithm for heterogeneous MTL is provided in the Supplementary Material \ref{app: algorithm}.

\section{Experiments}\label{sec: exp}
In this section, we conduct extensive numerical experiments, including two-task MTL, multi-task MTL with more than two tasks, and an application to the ADNI real dataset. The numerical experiments demonstrate that our proposed two-step MTL method  aggregates information in the presence of block-wise, distribution, and posterior heterogeneity.
\subsection{MTL for Two tasks}\label{sec: exp 2 tasks} 
We address a common real-world scenario involving MTL with two tasks for illustration. The data generation process (DGP) is as follows: For Task \num{1}, the features are denoted as $\vx^{1} = [\vx^{1}_{0}|\vx^{1}_{1}|\vx^{1}_{2}]$ and follow a Gaussian distribution with mean $\bm{0}$ and an exchangeable covariance matrix. The variance is fixed at $1$, and the covariance structure is specified as $(\rho_1)^{0.01 |i-j|}$. We randomly generate $n_1$ samples, with the third block $\vx^{1}_{2}$ missing in Task \num{1}. 
For Task \num{2}, the features are denoted as $\vx^2 = [\vx^{2}_{0}|\vx^{2}_{1}|\vx^{2}_{2}]$, where $\vx^2$ follows a Gaussian distribution with mean $\bm{0}$, variance $1$, and covariance structure $(\rho_2)^{0.01 |i-j|}$. In this task, we generate $n_2$ samples, with the second block $\vx^{2}_{1}$ missing.
For $t=1,2$, the responses are defined as:
\begin{align*}
y^{t} = \alpha \sum_{d=1}^{p} {v_{c,d}(x^{t}_d)^2}/{p} + (1-\alpha) \sum_{d=1}^{p} {v_{t,d}x^{t}_d}/{p} + \varepsilon_t,
\end{align*}
where $p = \sum_{s=0}^2 p_s$, with $p_s$ being the dimension of the $s$-th source, and the subscript $d$ denotes the $d$-th element of a vector (this notation applies to subsequent symbols as well). The parameters $v_c$, $v_1$, and $v_2$ are sampled from $N(-10, 10^2)$, and the noise terms $\varepsilon_1 \sim N(0, \sigma_1^2)$ and $\varepsilon_2 \sim N(0, \sigma_2^2)$. The parameter $\alpha$ controls the level of sharing across tasks. Additionally, our DGP accounts for \textbf{nonlinear relationships} by
element-wise square, further increasing the complexity of MTL.
For evaluation, we calculate the average root-mean-squared error (RMSE) on the testing data, as defined in the Supplementary Material \ref{App: DPG}. We conduct experiments under various settings to compare the proposed MTL for heterogeneous multi-source block-wise missing data (MTL-HMB) against existing methods, including Single Task Learning (STL) and Transfer Learning for Heterogeneous Data (HTL) \citep{bica2022transfer}. 

\begin{figure*}[t]
\centering
\subfigure[$\rho_1=\rho_2$]{\label{subfig:a}
\includegraphics[width=0.31\textwidth]{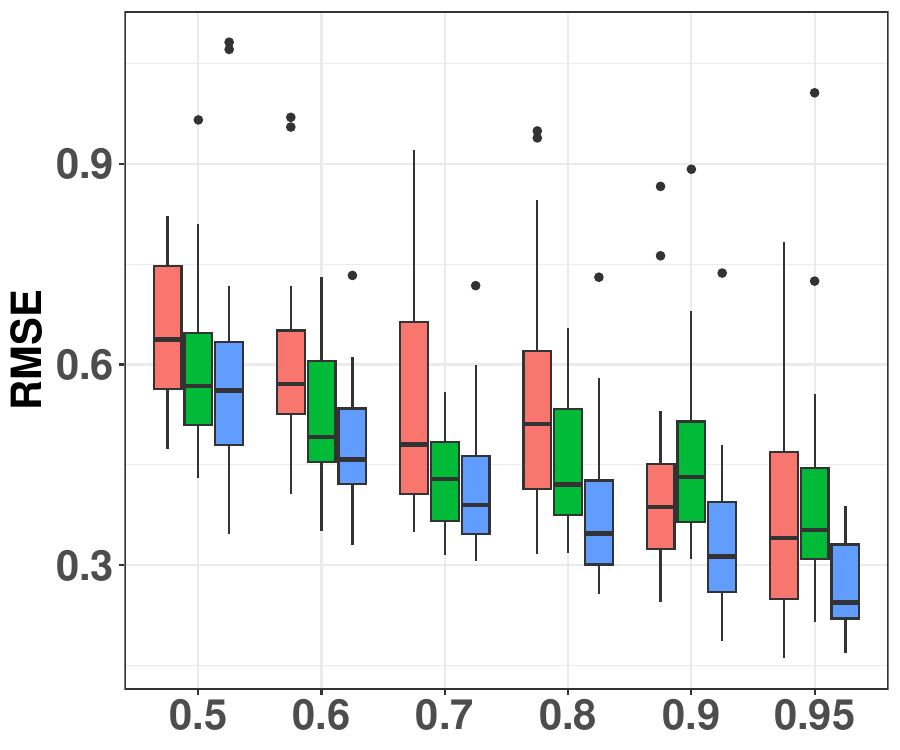}
}
\subfigure[$\rho_1\neq\rho_2$]{\label{subfig:b}
\includegraphics[width=0.31\textwidth]{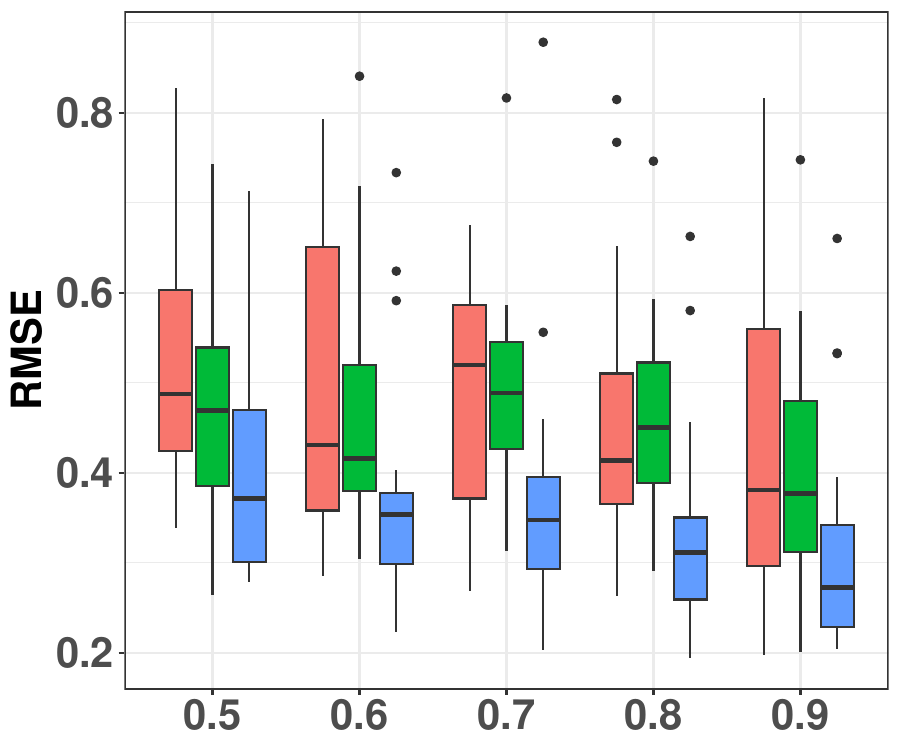}
}
\subfigure[$\alpha$]{\label{subfig:c}
\includegraphics[width=0.31\textwidth]{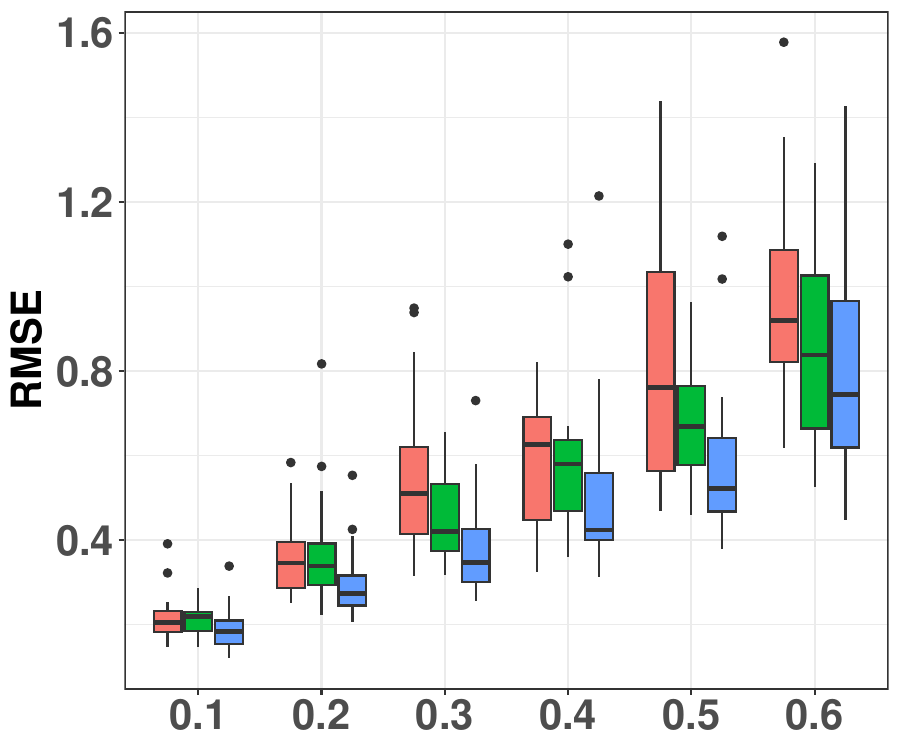}
}
\subfigure[$n_1=n_2$]{\label{subfig:d}
\includegraphics[width=0.31\textwidth]{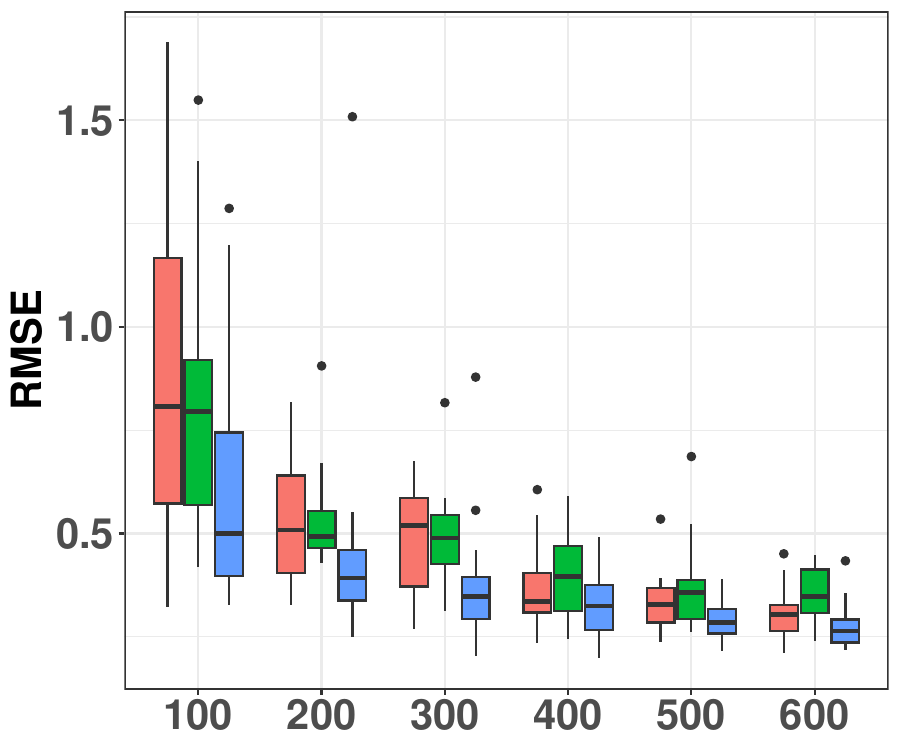}
}
\subfigure[$p_1=p_2$]{\label{subfig:e}
\includegraphics[width=0.31\textwidth]{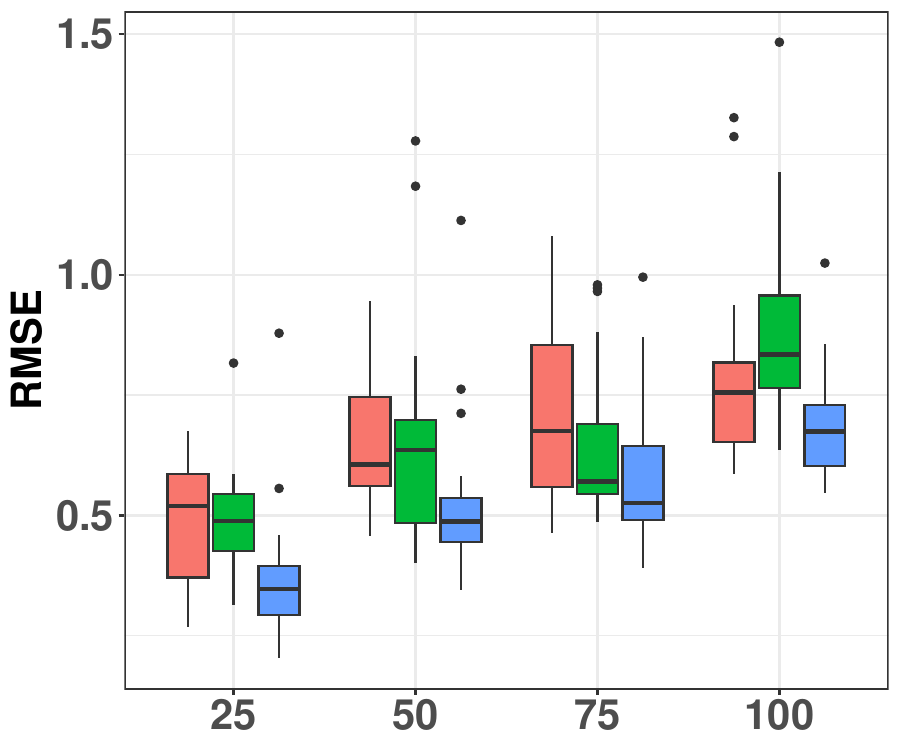}
}
\subfigure[$\sigma_1\neq\sigma_2$]{\label{subfig:f}
\includegraphics[width=0.31\textwidth]{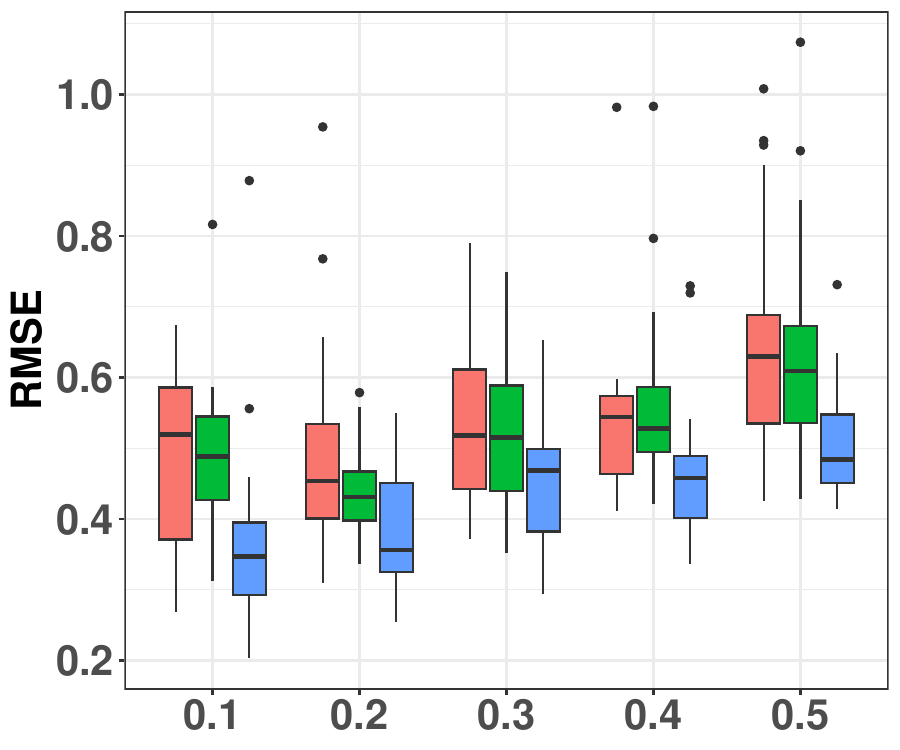}
}
\caption{Boxplots of average RMSEs under \textbf{Settings A} to \textbf{F} for the three methods. The methods are distinguished by color: \textcolor{ggplotred}{\textbf{orange}} for STL, \textcolor{ggplotgreen}{\textbf{green}} for HTL, and \textcolor{ggplotblue}{\textbf{blue}} for the proposed MTL-HMB.
}
\label{fig:initial}
\end{figure*}

\textbf{Setting A: Effect of covariance parameters.} We set $n_1 = n_2 = \num{300}$, $p_0 = 100$, $p_1 = p_2 = 25$, $\alpha = 0.3$, and $\sigma_1 = \sigma_2 = 0.1$. To examine the impact of the covariance parameters $\rho_1$ and $\rho_2$, we set $\rho_1 = \rho_2$ and vary them from $0.5$ to $0.95$, assuming no distribution heterogeneity across datasets. As shown in Figure \ref{subfig:a}, increasing correlation improves prediction accuracy across all methods. The proposed MTL-HMB is the best performer. Specifically, at $\rho = 0.95$, it outperforms the others by more than \SI{28.33}{\percent}. Even at $\rho = 0.5$, despite imputation errors, our method maintains an advantage. This demonstrates that imputation enhances prediction, especially when distribution heterogeneity is absent.

\textbf{Setting B: Effect of heterogeneous covariance parameters.} We set $n_1 = n_2 = \num{300}$, $p_0 = 100$, $p_1 = p_2 = 25$, $\alpha = 0.3$, and $\sigma_1 = \sigma_2 = 0.1$. To assess the impact of heterogeneous covariance, we fix $\rho_1 = 0.95$ and vary $\rho_2$ from $0.5$ to $0.9$. Smaller $\rho_2$ indicates greater heterogeneity and weaker correlations in Task \num{2}, making predictions more challenging. Figure \ref{subfig:b} shows that as $\rho_2$ increases, all methods improve, and our approach consistently leads. At the highest level of heterogeneity, MTL-HMB outperforms HTL by over \SI{20.91}{\percent}. Moreover, HTL shows no advantage over STL, indicating that transfer learning struggles with distribution heterogeneity. In contrast, the proposed method effectively solves the heterogeneity challenge through imputation, achieving better predictive accuracy.

\textbf{Setting C: Effect of heterogeneous mappings.} We set $n_1 = n_2 = \num{300}$, $p_0 = 100$, $p_1 = p_2 = 25$, $\rho_1 = \rho_2 = 0.8$, and $\sigma_1 = \sigma_2 = 0.1$. The parameter $\alpha$ is varied to control the level of information sharing in the mappings to the response. A larger $\alpha$ indicates more shared information. With $\rho_1 = \rho_2$ fixed, heterogeneity is governed solely by $\alpha$. Figure \ref{subfig:c} shows that as $\alpha$ increases, the magnitude of $y$ also increases, resulting in higher average RMSEs. Except at $\alpha = 0.1$, HTL consistently outperforms STL, indicating its advantage in incorporating posterior shift. Overall, the proposed MTL-HMB performs the best across all settings, even in the absence of distribution heterogeneity.

\textbf{Setting D: Effect of sample sizes.} We set $p_0 = 100$, $p_1 = p_2 = 25$, $\rho_1 = 0.95$, $\rho_2 = 0.7$, $\alpha = 0.3$, and $\sigma_1 = \sigma_2 = 0.1$. The sample sizes $n_1$ and $n_2$ vary as $n_1 = n_2 = k \times 100$ for $k = 1, \ldots, 6$. Figure \ref{subfig:d} shows that as the sample size increases, average RMSEs decrease, and the corresponding variability of estimator is reduced across all methods. Our method consistently performs best, with an improvement of over \SI{18.13}{\percent} compared to HTL. This is particularly notable at smaller sample sizes such as \num{100}, where MTL-HMB outperforms HTL and STL by \SI{37.13}{\percent} and \SI{38.06}{\percent}, respectively. Additionally, HTL does not significantly outperform STL, indicating difficulty in handling distribution heterogeneity.

\textbf{Setting E: Effect of dimensions.} We set $n_1 = n_2 = \num{300}$, $\rho_1 = 0.95$, $\rho_2 = 0.7$, $\alpha = 0.3$, and $\sigma_1 = \sigma_2 = 0.1$. To assess the impact of dimensions $p_1$, $p_2$, and $p_3$, we fix $p_1 = 100$ and vary $p_2 = p_3 = k \times \num{25}$ for $k = 1, \ldots, 4$. Figure \ref{subfig:e} shows that increasing dimensions make prediction more challenging, leading to higher average RMSEs for all methods. Our method consistently outperforms the others, with at least \SI{7.88}{\percent} and \SI{10.73}{\percent} improvements over HTL and STL, respectively. Moreover, it exhibits greater stability, refleced by lower RMSEs at both the \num{75}th and \num{25}th percentiles.

\textbf{Setting F: Effect of heterogeneous noise levels.} We set $n_1 = n_2 = \num{300}$, $p_0 = 100$, $p_1 = p_2 = 25$, $\rho_1 = 0.95$, $\rho_2 = 0.7$. By fixing $\sigma_2 = 0.1$ and varying $\sigma_1$ from \num{0.1} to \num{0.5}, we assess the impact of different noise levels. Figure \ref{subfig:f} shows that HTL lacks a clear advantage over STL, indicating that distribution heterogeneity leads to degenerated HTL's performance. Our MTL-HMB consistently outperforms the competing methods, demonstrating robustness in addressing both distribution and posterior heterogeneity. Furthermore, MTL-HMB excels at lower prediction levels, with neither STL nor HTL matching its performance at the \num{25}th percentile.
\subsection{MTL for Multiple tasks}\label{sec: exp multi}
\begin{figure*}[t]
\centering
\subfigure[Average]{\label{multi subfig:a}
\includegraphics[width=0.31\textwidth]{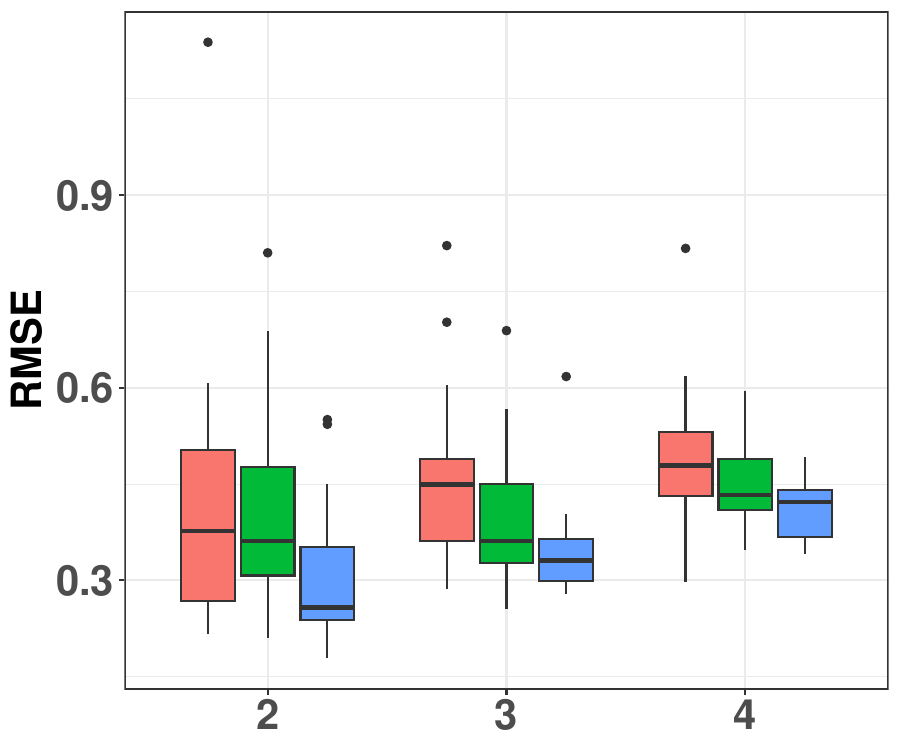}
}
\subfigure[\textquotedblleft Easy\textquotedblright]{\label{multi subfig:b}
\includegraphics[width=0.31\textwidth]{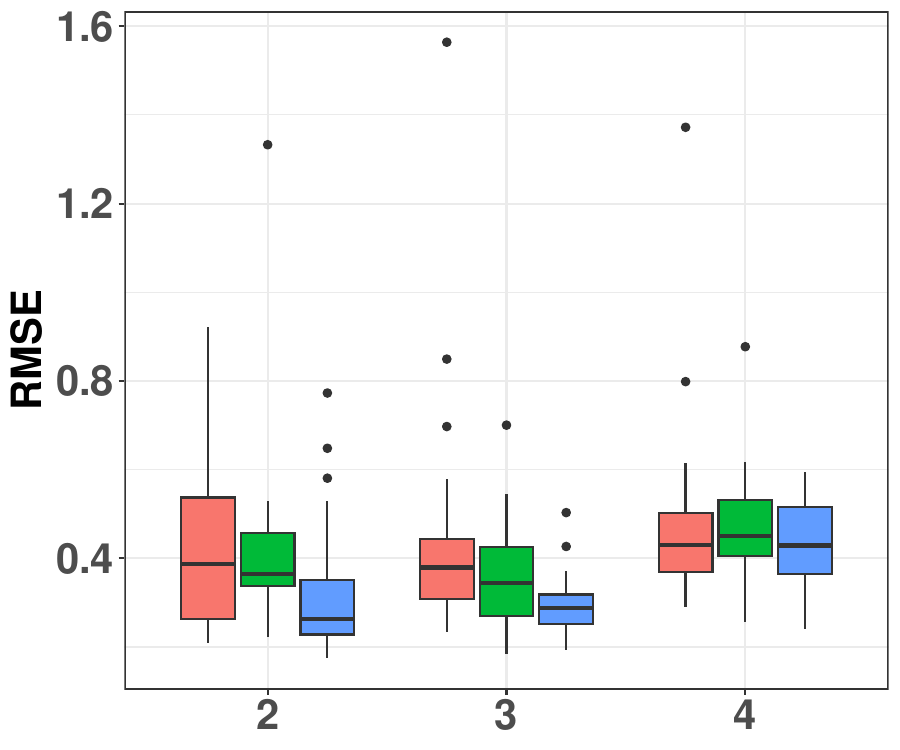}
}
\subfigure[\textquotedblleft Difficult\textquotedblright]{\label{multi subfig:c}
\includegraphics[width=0.31\textwidth]{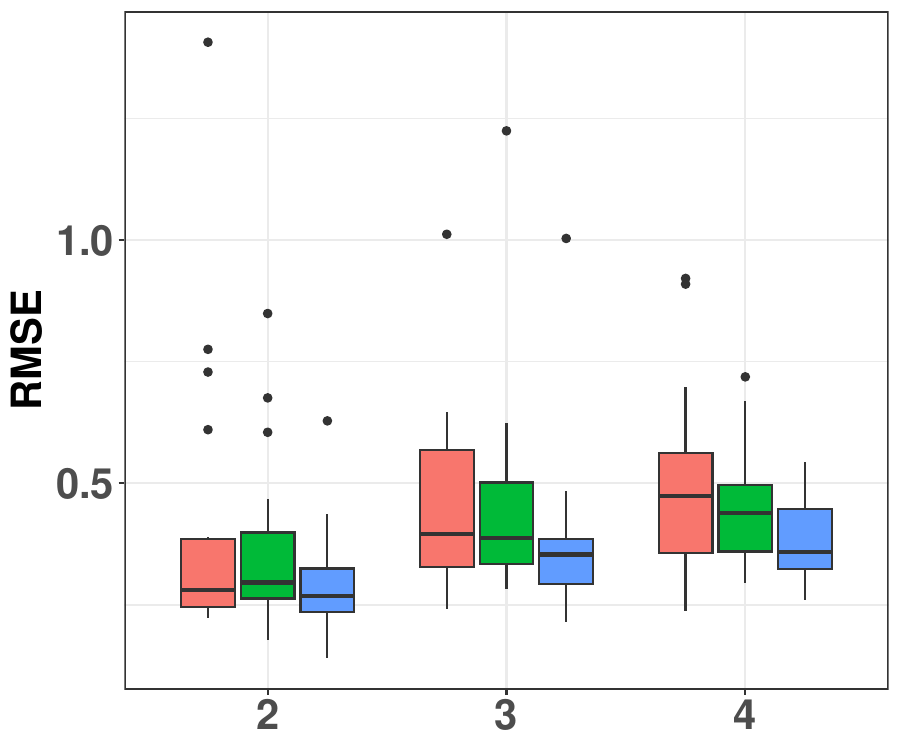}
}
\caption{Boxplots of average RMSEs under multiple settings across three methods. The methods are differentiated by color: \textcolor{ggplotred}{\textbf{orange}} for STL, \textcolor{ggplotgreen}{\textbf{green}} for HTL, and \textcolor{ggplotblue}{\textbf{blue}} for our proposed MTL-HMB.
}
\label{fig:multi}
\end{figure*}

To save space, the DGP for number of tasks greater than \num{2} is detailed in the Supplementary Material \ref{App: DPG}. We select $T = 2$, $T = 3$, and $T = 4$, with the prediction results summarized in Figure \ref{fig:multi}. As shown in Figure \ref{multi subfig:a}, increasing the number of heterogeneous tasks makes prediction more challenging, resulting in higher average RMSEs. This underscores the complexity of integrating diverse data. Nevertheless, our MTL-HMB consistently outperforms the other methods and shows the smallest RMSE standard deviation, indicating greater robustness and reliability. We focus on the first task, considering it the \textquotedblleft easiest\textquotedblright\ due to the highest observed correlations. Figure \ref{multi subfig:b} shows that for this \textquotedblleft easy\textquotedblright\ task, as more tasks are integrated, the improvement of our method decreases due to increasing heterogeneity. Additionally, we observe that in two-task learning, the second task is the most challenging; in three-task learning, it is the third task; and in four-task learning, it is the fourth task. This pattern indicates that as the number of integrated tasks increases, the complexity of learning escalates, particularly for the most recently added task. To quantify these challenges, we compile the RMSEs for the most difficult tasks in Figure \ref{multi subfig:c}, which shows that MTL-HMB excels in these challenging tasks, consistently outperforming the other methods. For example, in the four-task integration, our method achieves over \SI{18.22}{\percent} improvement compared to HTL.
\subsection{ADNI Real Data Application}\label{sec: adni}
We perform MTL using the ADNI database. The first task has \num{72} samples with features from MRI and PET sources, denoted as $\mX^{1}_{0}$ and $\mX^{1}_{1}$. The second task has \num{69} samples with features from MRI and GENE sources, denoted as $\mX^{2}_{0}$ and $\mX^{2}_{2}$. The MRI source includes \num{267} features, PET includes \num{113}, and GENE includes \num{300}. For the response variable, we use the Mini-Mental State Examination (MMSE), which measures cognitive impairment and serves as a diagnostic indicator of Alzheimer’s disease \citep{tombaugh1992mini}. We provide a detailed description of the ADNI database in the Supplementary Material \ref{sec: real data des}.

Although both tasks share the MRI source, significant heterogeneity may still exist between the two datasets. To quantitatively assess this heterogeneity, we calculate the Maximum Mean Discrepancy (MMD) distance between $\mX^1_0$ and $\mX^2_0$. Additionally, a permutation test is conducted to determine whether the differences between these sample sets are statistically significant. The test yields a $p$-value of $1 \times 10^{-6}$, indicating significant differences between $\mX^1_0$ and $\mX^2_0$ and therefore a necessity of incorporating heterogeneity among homogeneous source in MTL. Furthermore, the small sample sizes in both tasks impose challenges for prediction, where MTL can potentially enhance performance. 
For both datasets, we use \num{60}\% of the samples for training, \num{20}\% for model selection and early stopping, and calculate RMSE on the remaining \num{20}\% for testing. The experiment is repeated \num{30} times, and the results are summarized in Table \ref{tab:losses}. Our MTL-HMB yields lower prediction errors in both tasks, particularly in Task \num{2}, where it improves performance by at least \num{17.28}\% compared to the other two methods, despite the small sample sizes. HTL performs worse than STL due to ignoring heterogeneity between $\mX^1_{0}$ and $\mX^{2}_{0}$.

\begin{table}[h]
\caption{Prediction accuracy on testing data, measured by RMSE.}
\label{tab:losses}
\begin{center}
\begin{small}
\begin{tabular}{lll}
    \toprule
    Method & Task \num{1} & Task \num{2} \\
    \midrule
       STL & $2.74 \, (0.87)$ & $4.57 \, (1.15)$ \\
       HTL & $2.86 \, (0.75)$ & $4.34 \, (1.47)$ \\
       \textbf{Ours} & $\bf 2.66 \, (0.59)$ & $\bf 3.59 \, (0.98)$ \\
    \bottomrule
\end{tabular}
\end{small}
\end{center}
\end{table}

\begin{figure}[t]
\begin{center}
\centerline{\includegraphics[width=0.4\columnwidth]{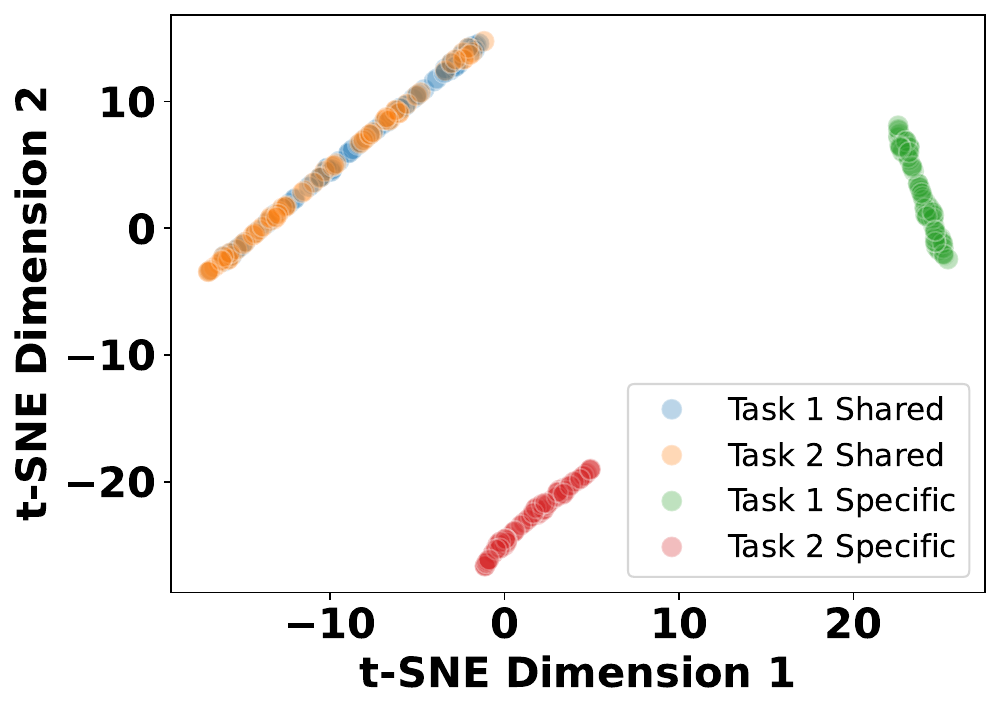}}
    \caption{The t-SNE visualization of the learned task-specific and shared
representations.}
    \label{fig:tsne}
\end{center}
\end{figure}

Figure \ref{fig:tsne} presents the t-SNE visualization of the latent representations obtained from a single training session, where the proposed MTL-HMB method effectively captures both shared and task-specific representations. Notably, the shared representations of the two tasks form a single cluster, while the task-specific representations of the two tasks exhibit significant differences in their distributions. This indicates that the datasets for the two tasks share certain commonalities while also displaying clear heterogeneity, which requires careful consideration during integration.
\section{Discussion}
In this paper, we propose a novel two-step strategy for effective MTL in the context of block-wise missing data in conjunction with different types of heterogeneity. The first step addresses distribution heterogeneity using integrated imputation, while the second step integrates learning to overcome distribution and posterior heterogeneity. We conduct extensive numerical experiments to validate the superiority of the proposed method across various levels of heterogeneity. Additionally, in the ADNI real-world dataset, our method achieves significant improvements in both tasks. 
In the following, we provide the limitations and outline future work directions, primarily focusing on transforming the two-step process into a single-step approach. In this unified method, the shared and task-specific hidden representations can be used for both imputing missing data and posterior learning simultaneously, as detailed in the Supplementary Material \ref{sec: future}.
\bibliographystyle{apalike}
\bibliography{conferences}

\begin{thebibliography}{}

\bibitem[Abdollahzadeh et~al., 2021]{abdollahzadeh2021revisit}
Abdollahzadeh, M., Malekzadeh, T., and Cheung, N.-M.~M. (2021).
\newblock Revisit multimodal meta-learning through the lens of multi-task learning.
\newblock {\em Advances in Neural Information Processing Systems}, 34:14632--14644.

\bibitem[Anceschi et~al., 2024]{anceschi2024bayesian}
Anceschi, N., Ferrari, F., Dunson, D.~B., and Mallick, H. (2024).
\newblock Bayesian joint additive factor models for multiview learning.
\newblock {\em arXiv preprint arXiv:2406.00778}.

\bibitem[Bai et~al., 2022]{bai2022contrastive}
Bai, T., Xiao, Y., Wu, B., Yang, G., Yu, H., and Nie, J.-Y. (2022).
\newblock A contrastive sharing model for multi-task recommendation.
\newblock In {\em Proceedings of the ACM Web Conference 2022}, pages 3239--3247.

\bibitem[Bernal et~al., 2019]{bernal2019deep}
Bernal, J., Kushibar, K., Asfaw, D.~S., Valverde, S., Oliver, A., Mart{\'\i}, R., and Llad{\'o}, X. (2019).
\newblock Deep convolutional neural networks for brain image analysis on magnetic resonance imaging: a review.
\newblock {\em Artificial intelligence in medicine}, 95:64--81.

\bibitem[Bica and van~der Schaar, 2022]{bica2022transfer}
Bica, I. and van~der Schaar, M. (2022).
\newblock Transfer learning on heterogeneous feature spaces for treatment effects estimation.
\newblock {\em Advances in Neural Information Processing Systems}, 35:37184--37198.

\bibitem[Bousmalis et~al., 2016]{bousmalis2016domain}
Bousmalis, K., Trigeorgis, G., Silberman, N., Krishnan, D., and Erhan, D. (2016).
\newblock Domain separation networks.
\newblock {\em Advances in neural information processing systems}, 29.

\bibitem[Cai et~al., 2024]{caibandits2024transfer}
Cai, C., Cai, T.~T., and Li, H. (2024).
\newblock Transfer learning for contextual multi-armed bandits.
\newblock {\em The Annals of Statistics}, 52(1):207--232.

\bibitem[Cai et~al., 2022]{cai2022individual}
Cai, T., Liu, M., and Xia, Y. (2022).
\newblock Individual data protected integrative regression analysis of high-dimensional heterogeneous data.
\newblock {\em Journal of the American Statistical Association}, 117(540):2105--2119.

\bibitem[Cai and Pu, 2024]{cai2024transfer}
Cai, T.~T. and Pu, H. (2024).
\newblock Transfer learning for nonparametric regression: Non-asymptotic minimax analysis and adaptive procedure.
\newblock {\em arXiv preprint arXiv:2401.12272}.

\bibitem[Cao et~al., 2022a]{cao2022unified}
Cao, K., Gong, Q., Hong, Y., and Wan, L. (2022a).
\newblock A unified computational framework for single-cell data integration with optimal transport.
\newblock {\em Nature Communications}, 13(1):7419.

\bibitem[Cao et~al., 2022b]{cao2022manifold}
Cao, K., Hong, Y., and Wan, L. (2022b).
\newblock Manifold alignment for heterogeneous single-cell multi-omics data integration using pamona.
\newblock {\em Bioinformatics}, 38(1):211--219.

\bibitem[Chaudhry et~al., 2018]{chaudhry2018efficient}
Chaudhry, A., Ranzato, M., Rohrbach, M., and Elhoseiny, M. (2018).
\newblock Efficient lifelong learning with a-gem.
\newblock {\em arXiv preprint arXiv:1812.00420}.

\bibitem[Chen et~al., 2021]{chen2021identifying}
Chen, J., Tran-Dinh, Q., Kosorok, M.~R., and Liu, Y. (2021).
\newblock Identifying heterogeneous effect using latent supervised clustering with adaptive fusion.
\newblock {\em Journal of Computational and Graphical Statistics}, 30(1):43--54.

\bibitem[Choi and Jung, 2022]{choi2022integrative}
Choi, S.~G. and Jung, S. (2022).
\newblock Integrative decomposition of multi-source data by identifying partially-joint score subspaces.
\newblock {\em arXiv preprint arXiv:2203.14041}.

\bibitem[Clark et~al., 2019]{clark2019bam}
Clark, K., Luong, M.-T., Khandelwal, U., Manning, C.~D., and Le, Q.~V. (2019).
\newblock Bam! born-again multi-task networks for natural language understanding.
\newblock {\em arXiv preprint arXiv:1907.04829}.

\bibitem[Cohen~Kalafut et~al., 2023]{cohen2023joint}
Cohen~Kalafut, N., Huang, X., and Wang, D. (2023).
\newblock Joint variational autoencoders for multimodal imputation and embedding.
\newblock {\em Nature Machine Intelligence}, 5(6):631--642.

\bibitem[Colnet et~al., 2024]{colnet2024causal}
Colnet, B., Mayer, I., Chen, G., Dieng, A., Li, R., Varoquaux, G., Vert, J.-P., Josse, J., and Yang, S. (2024).
\newblock Causal inference methods for combining randomized trials and observational studies: a review.
\newblock {\em Statistical science}, 39(1):165--191.

\bibitem[Crawshaw, 2020]{crawshaw2020multi}
Crawshaw, M. (2020).
\newblock Multi-task learning with deep neural networks: A survey.
\newblock {\em arXiv preprint arXiv:2009.09796}.

\bibitem[Curth and Van~der Schaar, 2021]{curth2021inductive}
Curth, A. and Van~der Schaar, M. (2021).
\newblock On inductive biases for heterogeneous treatment effect estimation.
\newblock {\em Advances in Neural Information Processing Systems}, 34:15883--15894.

\bibitem[D'Eramo et~al., 2024]{d2024sharing}
D'Eramo, C., Tateo, D., Bonarini, A., Restelli, M., and Peters, J. (2024).
\newblock Sharing knowledge in multi-task deep reinforcement learning.
\newblock {\em arXiv preprint arXiv:2401.09561}.

\bibitem[Du et~al., 2018]{du2018adapting}
Du, Y., Czarnecki, W.~M., Jayakumar, S.~M., Farajtabar, M., Pascanu, R., and Lakshminarayanan, B. (2018).
\newblock Adapting auxiliary losses using gradient similarity.
\newblock {\em arXiv preprint arXiv:1812.02224}.

\bibitem[Duan and Wang, 2023]{duan2023adaptive}
Duan, Y. and Wang, K. (2023).
\newblock Adaptive and robust multi-task learning.
\newblock {\em The Annals of Statistics}, 51(5):2015--2039.

\bibitem[Farahani et~al., 2021]{farahani2021brief}
Farahani, A., Voghoei, S., Rasheed, K., and Arabnia, H.~R. (2021).
\newblock A brief review of domain adaptation.
\newblock {\em Advances in data science and information engineering: proceedings from ICDATA 2020 and IKE 2020}, pages 877--894.

\bibitem[Feng et~al., 2018]{feng2018angle}
Feng, Q., Jiang, M., Hannig, J., and Marron, J. (2018).
\newblock Angle-based joint and individual variation explained.
\newblock {\em Journal of multivariate analysis}, 166:241--265.

\bibitem[Gao and Lee, 2017]{gao2017high}
Gao, Q. and Lee, T.~C. (2017).
\newblock High-dimensional variable selection in regression and classification with missing data.
\newblock {\em Signal Processing}, 131:1--7.

\bibitem[Gao et~al., 2021]{gao2021covariate}
Gao, X., Lee, S., Li, G., and Jung, S. (2021).
\newblock Covariate-driven factorization by thresholding for multiblock data.
\newblock {\em Biometrics}, 77(3):1011--1023.

\bibitem[Gaynanova and Li, 2019]{gaynanova2019structural}
Gaynanova, I. and Li, G. (2019).
\newblock Structural learning and integrative decomposition of multi-view data.
\newblock {\em Biometrics}, 75(4):1121--1132.

\bibitem[Gong et~al., 2019]{gong2019comparison}
Gong, T., Lee, T., Stephenson, C., Renduchintala, V., Padhy, S., Ndirango, A., Keskin, G., and Elibol, O.~H. (2019).
\newblock A comparison of loss weighting strategies for multi task learning in deep neural networks.
\newblock {\em IEEE Access}, 7:141627--141632.

\bibitem[Han et~al., 2024]{han2024parameter}
Han, Z., Gao, C., Liu, J., Zhang, S.~Q., et~al. (2024).
\newblock Parameter-efficient fine-tuning for large models: A comprehensive survey.
\newblock {\em arXiv preprint arXiv:2403.14608}.

\bibitem[Hang et~al., 2023]{hang2023efficient}
Hang, T., Gu, S., Li, C., Bao, J., Chen, D., Hu, H., Geng, X., and Guo, B. (2023).
\newblock Efficient diffusion training via min-snr weighting strategy.
\newblock In {\em Proceedings of the IEEE/CVF International Conference on Computer Vision}, pages 7441--7451.

\bibitem[He et~al., 2024a]{he2024transfusion}
He, Z., Sun, Y., and Li, R. (2024a).
\newblock Transfusion: Covariate-shift robust transfer learning for high-dimensional regression.
\newblock In {\em International Conference on Artificial Intelligence and Statistics}, pages 703--711. PMLR.

\bibitem[He et~al., 2024b]{he2024adatrans}
He, Z., Sun, Y., Liu, J., and Li, R. (2024b).
\newblock Adatrans: Feature-wise and sample-wise adaptive transfer learning for high-dimensional regression.
\newblock {\em arXiv preprint arXiv:2403.13565}.

\bibitem[Hu et~al., 2022]{hu2022improving}
Hu, Z., Zhao, Z., Yi, X., Yao, T., Hong, L., Sun, Y., and Chi, E. (2022).
\newblock Improving multi-task generalization via regularizing spurious correlation.
\newblock {\em Advances in Neural Information Processing Systems}, 35:11450--11466.

\bibitem[Huang et~al., 2023a]{huang2023integrative}
Huang, J., Jiao, Y., Wang, W., Yan, X., and Zhu, L. (2023a).
\newblock Integrative analysis for high-dimensional stratified models.
\newblock {\em Statistica Sinica}, 33.

\bibitem[Huang et~al., 2023b]{huangxin2023optimal}
Huang, X., Xu, K., Lee, D., Hassani, H., Bastani, H., and Dobriban, E. (2023b).
\newblock Optimal heterogeneous collaborative linear regression and contextual bandits.
\newblock {\em arXiv preprint arXiv:2306.06291}.

\bibitem[Imbens and Rubin, 2015]{imbens2015causal}
Imbens, G.~W. and Rubin, D.~B. (2015).
\newblock {\em Causal inference in statistics, social, and biomedical sciences}.
\newblock Cambridge university press.

\bibitem[James et~al., 2024]{james2024learning}
James, C., Yuan, D., Gaynanova, I., and Arroyo, J. (2024).
\newblock Learning joint and individual structure in network data with covariates.
\newblock {\em arXiv preprint arXiv:2406.08776}.

\bibitem[Jin et~al., 2017]{jin2017deep}
Jin, K.~H., McCann, M.~T., Froustey, E., and Unser, M. (2017).
\newblock Deep convolutional neural network for inverse problems in imaging.
\newblock {\em IEEE transactions on image processing}, 26(9):4509--4522.

\bibitem[Kent et~al., 2020]{kent2020predictive}
Kent, D.~M., Paulus, J.~K., Van~Klaveren, D., D'Agostino, R., Goodman, S., Hayward, R., Ioannidis, J.~P., Patrick-Lake, B., Morton, S., Pencina, M., et~al. (2020).
\newblock The predictive approaches to treatment effect heterogeneity (path) statement.
\newblock {\em Annals of internal medicine}, 172(1):35--45.

\bibitem[Kent et~al., 2018]{kent2018personalized}
Kent, D.~M., Steyerberg, E., and Van~Klaveren, D. (2018).
\newblock Personalized evidence based medicine: predictive approaches to heterogeneous treatment effects.
\newblock {\em Bmj}, 363.

\bibitem[Kouw and Loog, 2018]{kouw2018introduction}
Kouw, W.~M. and Loog, M. (2018).
\newblock An introduction to domain adaptation and transfer learning.
\newblock {\em arXiv preprint arXiv:1812.11806}.

\bibitem[Lam et~al., 2022]{lam2022adaptive}
Lam, H., Wang, K., Wu, Y., and Zhang, Y. (2022).
\newblock Adaptive data fusion for multi-task non-smooth optimization.
\newblock {\em arXiv preprint arXiv:2210.12334}.

\bibitem[Le~Morvan et~al., 2021]{le2021sa}
Le~Morvan, M., Josse, J., Scornet, E., and Varoquaux, G. (2021).
\newblock What’sa good imputation to predict with missing values?
\newblock {\em Advances in Neural Information Processing Systems}, 34:11530--11540.

\bibitem[Lee et~al., 2024]{lee2024enhancing}
Lee, H., Nam, G., Fong, E., and Lee, J. (2024).
\newblock Enhancing transfer learning with flexible nonparametric posterior sampling.
\newblock {\em arXiv preprint arXiv:2403.07282}.

\bibitem[Lee et~al., 2018]{lee2018joint}
Lee, J.-H., Chan, Y.-M., Chen, T.-Y., and Chen, C.-S. (2018).
\newblock Joint estimation of age and gender from unconstrained face images using lightweight multi-task cnn for mobile applications.
\newblock In {\em 2018 IEEE conference on multimedia information processing and retrieval (MIPR)}, pages 162--165. IEEE.

\bibitem[Li and Sang, 2019]{li2019spatial}
Li, F. and Sang, H. (2019).
\newblock Spatial homogeneity pursuit of regression coefficients for large datasets.
\newblock {\em Journal of the American Statistical Association}.

\bibitem[Li et~al., 2023]{li2023weighted}
Li, J., Zhang, Q., Chen, S., and Fang, K. (2023).
\newblock Weighted multiple blockwise imputation method for high-dimensional regression with blockwise missing data.
\newblock {\em Journal of Statistical Computation and Simulation}, 93(3):459--474.

\bibitem[Li and Li, 2022]{li2022integrative}
Li, Q. and Li, L. (2022).
\newblock Integrative factor regression and its inference for multimodal data analysis.
\newblock {\em Journal of the American Statistical Association}, 117(540):2207--2221.

\bibitem[Li et~al., 2022]{li2022transfer}
Li, S., Cai, T.~T., and Li, H. (2022).
\newblock Transfer learning for high-dimensional linear regression: Prediction, estimation and minimax optimality.
\newblock {\em Journal of the Royal Statistical Society Series B: Statistical Methodology}, 84(1):149--173.

\bibitem[Li et~al., 2024a]{li2024combining}
Li, T., Shi, C., Wen, Q., Sui, Y., Qin, Y., Lai, C., and Zhu, H. (2024a).
\newblock Combining experimental and historical data for policy evaluation.
\newblock {\em arXiv preprint arXiv:2406.00317}.

\bibitem[Li et~al., 2024b]{li2024adaptive}
Li, Y., Yang, X., Wei, Y., and Liu, M. (2024b).
\newblock Adaptive and efficient learning with blockwise missing and semi-supervised data.
\newblock {\em arXiv preprint arXiv:2405.18722}.

\bibitem[Lin et~al., 2023]{lin2023optimal}
Lin, C., Peng, J., Qin, Y., Li, Y., and Yang, Y. (2023).
\newblock Optimal integrating learning for split questionnaire design type data.
\newblock {\em Journal of Computational and Graphical Statistics}, 32(3):1009--1023.

\bibitem[Liu et~al., 2019]{liu2019multi}
Liu, X., He, P., Chen, W., and Gao, J. (2019).
\newblock Multi-task deep neural networks for natural language understanding.
\newblock In {\em Proceedings of the 57th Annual Meeting of the Association for Computational Linguistics}, pages 4487--4496.

\bibitem[Lock et~al., 2013]{lock2013joint}
Lock, E.~F., Hoadley, K.~A., Marron, J.~S., and Nobel, A.~B. (2013).
\newblock Joint and individual variation explained (jive) for integrated analysis of multiple data types.
\newblock {\em The annals of applied statistics}, 7(1):523.

\bibitem[Lopez-Paz and Ranzato, 2017]{lopez2017gradient}
Lopez-Paz, D. and Ranzato, M. (2017).
\newblock Gradient episodic memory for continual learning.
\newblock {\em Advances in neural information processing systems}, 30.

\bibitem[Ma and Huang, 2017]{ma2017concave}
Ma, S. and Huang, J. (2017).
\newblock A concave pairwise fusion approach to subgroup analysis.
\newblock {\em Journal of the American Statistical Association}, 112(517):410--423.

\bibitem[Madden et~al., 2016]{madden2016missing}
Madden, J.~M., Lakoma, M.~D., Rusinak, D., Lu, C.~Y., and Soumerai, S.~B. (2016).
\newblock Missing clinical and behavioral health data in a large electronic health record (ehr) system.
\newblock {\em Journal of the American Medical Informatics Association}, 23(6):1143--1149.

\bibitem[Maity et~al., 2024]{maity2024linear}
Maity, S., Dutta, D., Terhorst, J., Sun, Y., and Banerjee, M. (2024).
\newblock A linear adjustment-based approach to posterior drift in transfer learning.
\newblock {\em Biometrika}, 111(1):31--50.

\bibitem[Maninis et~al., 2019]{maninis2019attentive}
Maninis, K.-K., Radosavovic, I., and Kokkinos, I. (2019).
\newblock Attentive single-tasking of multiple tasks.
\newblock In {\em Proceedings of the IEEE/CVF conference on computer vision and pattern recognition}, pages 1851--1860.

\bibitem[Mansour et~al., 2008]{mansour2008domain}
Mansour, Y., Mohri, M., and Rostamizadeh, A. (2008).
\newblock Domain adaptation with multiple sources.
\newblock {\em Advances in neural information processing systems}, 21.

\bibitem[Maziarka et~al., 2022]{maziarka2022relationship}
Maziarka, {\L}., Nowak, A., Wo{\l}czyk, M., and Bedychaj, A. (2022).
\newblock On the relationship between disentanglement and multi-task learning.
\newblock In {\em Joint European Conference on Machine Learning and Knowledge Discovery in Databases}, pages 625--641. Springer.

\bibitem[Meinshausen and B{\"u}hlmann, 2015]{meinshausen2015maximin}
Meinshausen, N. and B{\"u}hlmann, P. (2015).
\newblock Maximin effects in inhomogeneous large-scale data.
\newblock {\em The Annals of Statistics}, 43(4):1801--1830.

\bibitem[Moon and Carbonell, 2017]{moon2017completely}
Moon, S. and Carbonell, J.~G. (2017).
\newblock Completely heterogeneous transfer learning with attention-what and what not to transfer.
\newblock In {\em IJCAI}, volume~1, pages 1--2.

\bibitem[Mueller et~al., 2005a]{mueller2005alzheimer}
Mueller, S.~G., Weiner, M.~W., Thal, L.~J., Petersen, R.~C., Jack, C., Jagust, W., Trojanowski, J.~Q., Toga, A.~W., and Beckett, L. (2005a).
\newblock The alzheimer’s disease neuroimaging initiative.
\newblock {\em Neuroimaging Clinics of North America}, 15(4):869.

\bibitem[Mueller et~al., 2005b]{mueller2005ways}
Mueller, S.~G., Weiner, M.~W., Thal, L.~J., Petersen, R.~C., Jack, C.~R., Jagust, W., Trojanowski, J.~Q., Toga, A.~W., and Beckett, L. (2005b).
\newblock Ways toward an early diagnosis in alzheimer’s disease: the alzheimer’s disease neuroimaging initiative (adni).
\newblock {\em Alzheimer's \& Dementia}, 1(1):55--66.

\bibitem[Nair et~al., 2019]{nair2019covariate}
Nair, N.~G., Satpathy, P., Christopher, J., et~al. (2019).
\newblock Covariate shift: A review and analysis on classifiers.
\newblock In {\em 2019 Global Conference for Advancement in Technology (GCAT)}, pages 1--6. IEEE.

\bibitem[Ouyang et~al., 2024]{ouyang2024imputation}
Ouyang, Z., Wang, L., and Initiative†, A. D.~N. (2024).
\newblock Imputation-based variable selection method for block-wise missing data when integrating multiple longitudinal studies.
\newblock {\em Mathematics}, 12(7):951.

\bibitem[Palzer et~al., 2022]{palzer2022sjive}
Palzer, E.~F., Wendt, C.~H., Bowler, R.~P., Hersh, C.~P., Safo, S.~E., and Lock, E.~F. (2022).
\newblock sjive: Supervised joint and individual variation explained.
\newblock {\em Computational statistics \& data analysis}, 175:107547.

\bibitem[Patterson and Gibson, 2017]{patterson2017deep}
Patterson, J. and Gibson, A. (2017).
\newblock {\em Deep learning: A practitioner's approach}.
\newblock " O'Reilly Media, Inc.".

\bibitem[Ruder et~al., 2019]{ruder2019latent}
Ruder, S., Bingel, J., Augenstein, I., and S{\o}gaard, A. (2019).
\newblock Latent multi-task architecture learning.
\newblock In {\em Proceedings of the AAAI conference on artificial intelligence}, volume~33, pages 4822--4829.

\bibitem[Rusu et~al., 2015]{rusu2015policy}
Rusu, A.~A., Colmenarejo, S.~G., Gulcehre, C., Desjardins, G., Kirkpatrick, J., Pascanu, R., Mnih, V., Kavukcuoglu, K., and Hadsell, R. (2015).
\newblock Policy distillation.
\newblock {\em arXiv preprint arXiv:1511.06295}.

\bibitem[Shu et~al., 2020]{shu2020d}
Shu, H., Wang, X., and Zhu, H. (2020).
\newblock D-cca: A decomposition-based canonical correlation analysis for high-dimensional datasets.
\newblock {\em Journal of the American Statistical Association}, 115(529):292--306.

\bibitem[Song et~al., 2024]{song2024semi}
Song, S., Lin, Y., and Zhou, Y. (2024).
\newblock Semi-supervised inference for block-wise missing data without imputation.
\newblock {\em Journal of Machine Learning Research}, 25(99):1--36.

\bibitem[Subramanian et~al., 2020]{subramanian2020multi}
Subramanian, I., Verma, S., Kumar, S., Jere, A., and Anamika, K. (2020).
\newblock Multi-omics data integration, interpretation, and its application.
\newblock {\em Bioinformatics and biology insights}, 14:1177932219899051.

\bibitem[Subramanian et~al., 2018]{subramanian2018learning}
Subramanian, S., Trischler, A., Bengio, Y., and Pal, C.~J. (2018).
\newblock Learning general purpose distributed sentence representations via large scale multi-task learning.
\newblock In {\em International Conference on Learning Representations}.

\bibitem[Tang and Song, 2016]{tang2016fused}
Tang, L. and Song, P.~X. (2016).
\newblock Fused lasso approach in regression coefficients clustering--learning parameter heterogeneity in data integration.
\newblock {\em Journal of Machine Learning Research}, 17(113):1--23.

\bibitem[Tang et~al., 2021]{tang2021individualized}
Tang, X., Xue, F., and Qu, A. (2021).
\newblock Individualized multidirectional variable selection.
\newblock {\em Journal of the American Statistical Association}, 116(535):1280--1296.

\bibitem[Teh et~al., 2017]{teh2017distral}
Teh, Y., Bapst, V., Czarnecki, W.~M., Quan, J., Kirkpatrick, J., Hadsell, R., Heess, N., and Pascanu, R. (2017).
\newblock Distral: Robust multitask reinforcement learning.
\newblock {\em Advances in neural information processing systems}, 30.

\bibitem[Tian and Feng, 2023]{tian2023transfer}
Tian, Y. and Feng, Y. (2023).
\newblock Transfer learning under high-dimensional generalized linear models.
\newblock {\em Journal of the American Statistical Association}, 118(544):2684--2697.

\bibitem[Tian et~al., 2023]{tian2023learning}
Tian, Y., Gu, Y., and Feng, Y. (2023).
\newblock Learning from similar linear representations: Adaptivity, minimaxity, and robustness.
\newblock {\em arXiv preprint arXiv:2303.17765}.

\bibitem[Tian et~al., 2022]{tian2022unsupervised}
Tian, Y., Weng, H., and Feng, Y. (2022).
\newblock Unsupervised multi-task and transfer learning on gaussian mixture models.
\newblock {\em arXiv preprint arXiv:2209.15224}.

\bibitem[Tombaugh and McIntyre, 1992]{tombaugh1992mini}
Tombaugh, T.~N. and McIntyre, N.~J. (1992).
\newblock The mini-mental state examination: a comprehensive review.
\newblock {\em Journal of the American Geriatrics Society}, 40(9):922--935.

\bibitem[Tu et~al., 2022]{tu2022cross}
Tu, X., Cao, Z.-J., Mostafavi, S., Gao, G., et~al. (2022).
\newblock Cross-linked unified embedding for cross-modality representation learning.
\newblock {\em Advances in Neural Information Processing Systems}, 35:15942--15955.

\bibitem[Tzeng et~al., 2017]{tzeng2017adversarial}
Tzeng, E., Hoffman, J., Saenko, K., and Darrell, T. (2017).
\newblock Adversarial discriminative domain adaptation.
\newblock In {\em Proceedings of the IEEE conference on computer vision and pattern recognition}, pages 7167--7176.

\bibitem[Ullrich et~al., 2017]{ullrich2017soft}
Ullrich, K., Meeds, E., and Welling, M. (2017).
\newblock Soft weight-sharing for neural network compression.
\newblock {\em arXiv preprint arXiv:1702.04008}.

\bibitem[Wang et~al., 2023a]{wang2023minimax}
Wang, C., Wang, C., He, X., and Feng, X. (2023a).
\newblock Minimax optimal transfer learning for kernel-based nonparametric regression.
\newblock {\em arXiv preprint arXiv:2310.13966}.

\bibitem[Wang et~al., 2024a]{wang2024multi}
Wang, H., Li, Q., and Liu, Y. (2024a).
\newblock Multi-response regression for block-missing multi-modal data without imputation.
\newblock {\em Statistica Sinica}, 34(2):527.

\bibitem[Wang and Lock, 2024]{wang2024multiple}
Wang, J. and Lock, E.~F. (2024).
\newblock Multiple augmented reduced rank regression for pan-cancer analysis.
\newblock {\em Biometrics}, 80(1):ujad002.

\bibitem[Wang et~al., 2024b]{wang2024comprehensive}
Wang, L., Zhang, X., Su, H., and Zhu, J. (2024b).
\newblock A comprehensive survey of continual learning: theory, method and application.
\newblock {\em IEEE Transactions on Pattern Analysis and Machine Intelligence}.

\bibitem[Wang et~al., 2023b]{wang2023high}
Wang, P., Li, Q., Shen, D., and Liu, Y. (2023b).
\newblock High-dimensional factor regression for heterogeneous subpopulations.
\newblock {\em Statistica Sinica}, 33(1):27.

\bibitem[Wang et~al., 2018]{wang2018flexible}
Wang, P., Liu, Y., and Shen, D. (2018).
\newblock Flexible locally weighted penalized regression with applications on prediction of alzheimer’s disease neuroimaging initiative’s clinical scores.
\newblock {\em IEEE transactions on medical imaging}, 38(6):1398--1408.

\bibitem[Wang et~al., 2023c]{wang2023joint}
Wang, P., Wang, H., Li, Q., Shen, D., and Liu, Y. (2023c).
\newblock Joint and individual component regression.
\newblock {\em Journal of Computational and Graphical Statistics}, pages 1--16.

\bibitem[Wu et~al., 2024]{wu2024adaptive}
Wu, H., Li, B., Tian, L., Feng, J., and Dong, C. (2024).
\newblock An adaptive loss weighting multi-task network with attention-guide proposal generation for small size defect inspection.
\newblock {\em The Visual Computer}, 40(2):681--698.

\bibitem[Xiang et~al., 2014]{xiang2014bi}
Xiang, S., Yuan, L., Fan, W., Wang, Y., Thompson, P.~M., Ye, J., Initiative, A. D.~N., et~al. (2014).
\newblock Bi-level multi-source learning for heterogeneous block-wise missing data.
\newblock {\em NeuroImage}, 102:192--206.

\bibitem[Xue et~al., 2021]{xue2021statistical}
Xue, F., Ma, R., and Li, H. (2021).
\newblock Statistical inference for high-dimensional linear regression with blockwise missing data.
\newblock {\em arXiv preprint arXiv:2106.03344}.

\bibitem[Xue and Qu, 2021]{xue2021integrating}
Xue, F. and Qu, A. (2021).
\newblock Integrating multisource block-wise missing data in model selection.
\newblock {\em Journal of the American Statistical Association}, 116(536):1914--1927.

\bibitem[Yi et~al., 2023]{yi2023hierarchical}
Yi, S., Wong, R. K.~W., and Gaynanova, I. (2023).
\newblock Hierarchical nuclear norm penalization for multi-view data integration.
\newblock {\em Biometrics}, 79(4):2933--2946.

\bibitem[Yu et~al., 2020]{yu2020optimal}
Yu, G., Li, Q., Shen, D., and Liu, Y. (2020).
\newblock Optimal sparse linear prediction for block-missing multi-modality data without imputation.
\newblock {\em Journal of the American Statistical Association}, 115(531):1406--1419.

\bibitem[Yuan et~al., 2012]{yuan2012multi}
Yuan, L., Wang, Y., Thompson, P.~M., Narayan, V.~A., Ye, J., Initiative, A. D.~N., et~al. (2012).
\newblock Multi-source feature learning for joint analysis of incomplete multiple heterogeneous neuroimaging data.
\newblock {\em NeuroImage}, 61(3):622--632.

\bibitem[Zhang et~al., 2024a]{zhang2024concert}
Zhang, R., Zhang, Y., Qu, A., Zhu, Z., and Shen, J. (2024a).
\newblock Concert: Covariate-elaborated robust local information transfer with conditional spike-and-slab prior.
\newblock {\em arXiv preprint arXiv:2404.03764}.

\bibitem[Zhang et~al., 2024b]{zhang2024learning}
Zhang, X., Liu, J., and Zhu, Z. (2024b).
\newblock Learning coefficient heterogeneity over networks: A distributed spanning-tree-based fused-lasso regression.
\newblock {\em Journal of the American Statistical Association}, 119(545):485--497.

\bibitem[Zhang and Gaynanova, 2022]{zhang2022joint}
Zhang, Y. and Gaynanova, I. (2022).
\newblock Joint association and classification analysis of multi-view data.
\newblock {\em Biometrics}, 78(4):1614--1625.

\bibitem[Zhang and Yang, 2018]{zhang2018overview}
Zhang, Y. and Yang, Q. (2018).
\newblock An overview of multi-task learning.
\newblock {\em National Science Review}, 5(1):30--43.

\bibitem[Zhang and Yang, 2021]{zhang2021survey}
Zhang, Y. and Yang, Q. (2021).
\newblock A survey on multi-task learning.
\newblock {\em IEEE transactions on knowledge and data engineering}, 34(12):5586--5609.

\bibitem[Zhang and Zhu, 2022]{zhang2022transfer}
Zhang, Y. and Zhu, Z. (2022).
\newblock Transfer learning for high-dimensional quantile regression via convolution smoothing.
\newblock {\em arXiv preprint arXiv:2212.00428}.

\bibitem[Zhao et~al., 2016]{zhao2016partially}
Zhao, T., Cheng, G., and Liu, H. (2016).
\newblock A partially linear framework for massive heterogeneous data.
\newblock {\em Annals of statistics}, 44(4):1400.

\bibitem[Zhou et~al., 2021]{zhou2021multi}
Zhou, D., Cai, T., and Lu, J. (2021).
\newblock Multi-source learning via completion of block-wise overlapping noisy matrices.
\newblock {\em arXiv preprint arXiv:2105.10360}.

\bibitem[Zhou et~al., 2023]{zhou2023multi}
Zhou, D., Cai, T., and Lu, J. (2023).
\newblock Multi-source learning via completion of block-wise overlapping noisy matrices.
\newblock {\em Journal of Machine Learning Research}, 24(221):1--43.

\end{thebibliography}
\newpage
\section*{Supplementary Materials}\label{supp}
\renewcommand{\thesection}{S}
\renewcommand{\thesubsection}{\thesection.\arabic{subsection}}
\setcounter{section}{0} 
\setcounter{subsection}{0}
In the Supplementary Material \ref{A1}, we further expand on the related works described in Section \ref{sec: related work}. 

\noindent In the Supplementary Material \ref{App: DPG}, we provide the detailed DGP and numerical results in Section \ref{sec: exp multi}.

\noindent In the Supplementary Material \ref{A3}, we conduct ablation experiments to demonstrate the individual roles of the two steps in our proposed method.

\noindent In the Supplementary Material \ref{A7}, we compare with additional statistical MBI method.

\noindent In the Supplementary Material \ref{app: algorithm}, we provide the pseudo-code for the proposed MTL-HMB.

\noindent In the Supplementary Material \ref{A5}, we include detailed experimental information, including the real data description and implementation details.

\noindent In the Supplementary Material \ref{sec: future}, we discuss the limitations of our work and potential future research directions.
\subsection{Expanded related works}\label{A1}
\textbf{Multi-group data integration.}
Multi-group data integration and MTL share the common goal of learning from multiple datasets or tasks simultaneously. The input features and response of a single task can be viewed as a separate group. There are several existing methods in the statistical literature for multi-group data analysis, which can be broadly classified into three categories.
The first category designs specialized regression models \citep{meinshausen2015maximin, zhao2016partially, wang2018flexible, huang2023integrative, huangxin2023optimal} or factor regression models \citep{wang2023high, wang2023joint} to handle large-scale heterogeneous data and identify group-specific structures. The second category employs specified parameter space constraints, such as fused penalties, to estimate regression coefficients that capture subgroup structures \citep{tang2016fused, ma2017concave, chen2021identifying, li2019spatial, tang2021individualized, lam2022adaptive, duan2023adaptive, zhang2024learning}. The third category involves transfer learning, which borrows information from source data to target data \citep{li2022transfer, tian2022unsupervised, zhang2022transfer, tian2023transfer, wang2023minimax,cai2024transfer, caibandits2024transfer, he2024transfusion, zhang2024concert}.
The aforementioned multi-group data integration approaches address distribution and posterior heterogeneity but overlook block-wise missing issues. Additionally, most methods rely on structured model assumptions, such as linearity, limiting their capacity to capture complex relationships.

\textbf{Heterogeneous feature spaces.} Existing transfer learning methods mainly addressed either distribution shift or posterior shift separately, with fewer studies considering both types of shifts simultaneously. For instance, \citet{moon2017completely} investigated scenarios with both heterogeneous feature and label spaces in the context of natural language processing. They proposed a method that learned a common embedding for the features and labels and then established a mapping between them. Similarly, \citet{bica2022transfer} focused on a shared label space but assumed that all tasks had a common source, utilizing the same encoder to extract shared representations. However, this assumption was often unrealistic in practice. Even when  sources were identical, different tasks could exhibit significant heterogeneity due to variations in subjects, locations, and experimental settings. For example, in our ADNI real data (Section \ref{sec: adni}), tasks sharing MRI features might still differ due to varying experimental conditions.
A key distinction in our method is that we treat this problem as a block-wise missing data issue rather than simply considering each task to have only the observed features. This perspective aligns more closely with the reality of medical data, where missing problem is common, and these missing features can also influence the response.
Additionally, we focus on MTL, which is designed for numerous small-sample and challenging tasks. In contrast, transfer learning often assumed the existence of a large-scale dataset to support a smaller-sample task. For example, in the experiments conducted by \citet{bica2022transfer}, the source domain's sample size was typically more than ten times that of the target domain. However, in real-world scenarios, it is more common for all tasks to have relatively small and limited sample sizes. Our method aims to provide a more comprehensive and robust learning framework by integrating heterogeneous information across these small-sample tasks.

\textbf{Block-wise Statistical Methods.} Numerous statistical methods for block-wise missing data have been proposed, and we provide a more detailed discussion here. \citet{yu2020optimal,wang2024multi} learn linear predictors through covariance matrix and cross-covariance vector, which can be estimated with block-missing data without imputation. \citet{xue2021integrating,xue2021statistical} propose a multiple block-wise imputation (MBI) approach to construct estimating equations based on all available information and integrate estimating functions to achieve efficient estimators. \citet{li2024adaptive} leverages block-wise missing labeled samples and further enhances estimation efficiency by incorporating large unlabeled samples through imputation and projection. Their method is robust to model misspecification on the missing covariates. \citet{song2024semi} address a similar problem under the semi-supervised learning setting, employing a double debiased procedure without relying on imputation. \citet{zhou2023multi} develop an efficient block-wise overlapping noisy matrix integration algorithm to obtain multi-source embeddings. These methods have demonstrated strong performance in various real-world applications. For instance, \citet{zhou2023multi,li2024adaptive} validated their methods on electronic health record (EHR) data, demonstrating their effectiveness in real-world applications.
However, all the aforementioned methods suffer from several limitations. First, they primarily capture linear relationships and struggle to effectively learn nonlinear patterns. Many real-world datasets, such as multi-modal single-cell data \citep{tu2022cross,cohen2023joint} and imaging data \citep{jin2017deep,bernal2019deep}, exhibit complexities that further limit the applicability of these methods. This limitation underscores the motivation for adopting an encoder-decoder framework in our work. Second, these methods consider the homogeneous model setup for different tasks, for instance, assuming the same regression coefficients are applied to all tasks. However, data heterogeneity across tasks or sources are ubiquitous in real applications, either marginal distribution of sources or conditional distribution among sources can be distinct, which complicates the modeling procedure. This is another motivation of our project, to effectively handle multiple types of heterogeneity simultaneously.

\subsection{Data Generation Process and Results in Section \ref{sec: exp multi}}\label{App: DPG}
We consider MTL for multiple tasks. The DGP is similar to that in Section \ref{sec: exp 2 tasks} but is extended to accommodate more tasks.
For three-task learning, the features for the $t$-th task are denoted as $\vx^t = [\vx^t_0|\vx^t_1|\vx^t_2|\vx^t_3]$ and follow a Gaussian distribution with mean $\bm{0}$ and an exchangeable covariance matrix. The variance is fixed at $1$, and the covariance structure is determined by $(\rho_t)^{0.01 |i-j|}$. We randomly generate $n_t$ samples, with only $\vx^t_0$ and $\vx^t_t$ being observed. The response $y^t$ is given by:
\begin{align*}
  y^t={\alpha}\sum_{d=1}^{p}v_{c,d}(x^t_d)^2/p+(1-\alpha)\sum_{d=1}^{p}v_{t,d}x^t_d/p+\varepsilon_r,\quad \forall r\in[3]. 
\end{align*}
where $p = \sum_{s=0}^3 p_s$. For three-task learning, we choose the following parameters: $n_1 = n_2 = n_3 = 300$, $p_0 = 125$, $p_1 = p_2 = p_3 = 25$, $\rho_1 = 0.95$, $\rho_2 = \rho_3 = 0.9$, $\alpha = 0.3$, $v_c, v_t \sim N(-10, 10^2)$, and $\varepsilon_t \sim N(0, 0.01)$ for $t \in [3]$.

For four-task learning, the features for the $t$-th task are denoted as $\vx^t = [\vx^t_0|\vx^t_1|\vx^t_2|\vx^t_3|\vx^t_4]$ and follow a Gaussian distribution with mean $\bm{0}$ and an exchangeable covariance matrix. The variance is fixed at $1$, and the covariance structure is determined by $(\rho_t)^{0.01 |i-j|}$. We randomly generate $n_t$ samples, with only $\vx^t_0$ and $\vx^t_t$ being observed. The response $y^t$ is given by:
\begin{align*}
  y^t={\alpha}\sum_{d=1}^{p}v_{c,d}(x^t_d)^2/p+(1-\alpha)\sum_{d=1}^{p}v_{t,d}x^t_d/p+\varepsilon_r,\quad \forall r\in[4]. 
\end{align*}
where $p = \sum_{s=0}^4 p_s$. For four-task learning, we choose the following parameters: $n_1 = n_2 = n_3 = n_4 = 300$, $p_0 = 125$, $p_1 = p_2 = p_3 = p_4 = 25$, $\rho_1 = 0.95$, $\rho_2 = \rho_3 = \rho_4 = 0.9$, $\alpha = 0.3$, $v_c, v_t \sim N(-10, 10^2)$, and $\varepsilon_t \sim N(0, 0.01)$ for $t \in [4]$. 

For evaluation, we focus on the average RMSE across all tasks in the testing data, defined as follows:
$$
\text{RMSE} = \frac{1}{T} \sum_{t=1}^T \sqrt{\frac{1}{n_{t, \text{test}}} \sum_{i=1}^{n_{t, \text{test}}} \left(\widehat{y}^t_i - y^t_i\right)^2}.
$$

\subsection{Ablation Experiments}\label{A3}
We propose an MTL framework that involves two steps: Step \num{1} for HBI (see Section \ref{sec: imputation}) and Step \num{2} for heterogeneous MTL (see Section \ref{sec: integration}). To assess the independent effect of each step, we design ablation experiments. In addition to comparing with STL and HTL, we consider three new ablation experiments. The first is Step \num{1} + STL, which applies HBI followed by STL to evaluate the effect of Step \num{2} and is denoted as Ablation \num{1}. The second approach is Step \num{1} combined with a common MTL framework (Ablation \num{2}). Specifically, we adopt the hard parameter sharing (HPS) framework, which shares the main layers across tasks while differentiating in the final layer, and is widely used in MTL \citep{liu2019multi, bai2022contrastive}. However, hard parameter sharing struggles to address distribution heterogeneity due to the shared structure in the first $L-1$ layers.
The third is naive imputation + Step \num{2}, where we ignore distribution heterogeneity to analyze the impact of disregarding heterogeneity in imputation, denoted as Ablation \num{3}.

\begin{figure}
    \centering
    \includegraphics[width=0.7\linewidth]{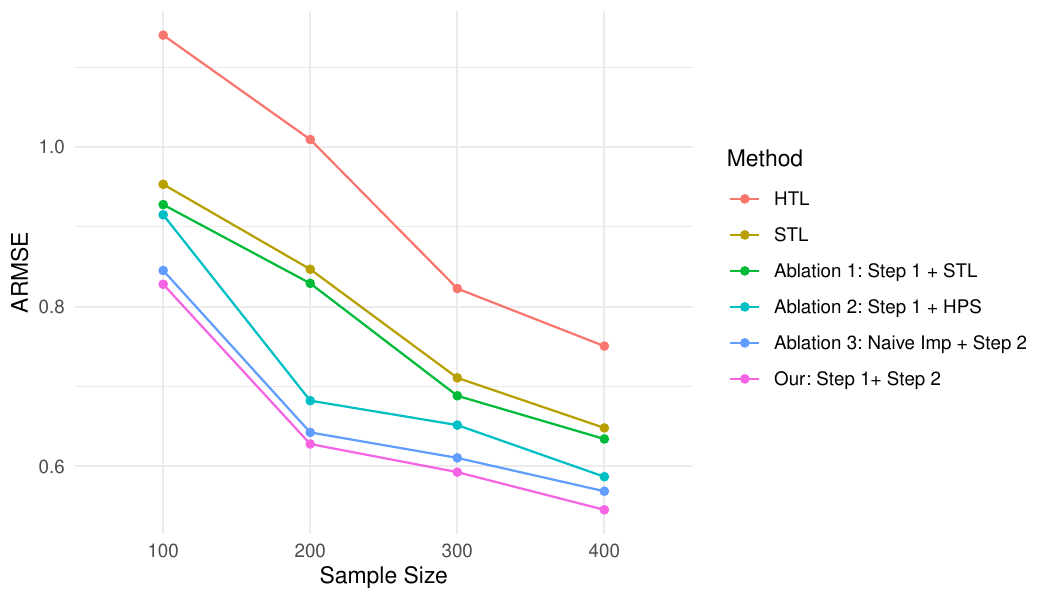}
    \caption{The average RMSEs of all methods across different $n_2$ sample sizes. HPS refers to hard parameter sharing, and Imp refers to imputation.}
    \label{fig:ablation}
\end{figure}

The data generation process is consistent with Section \ref{sec: exp 2 tasks}, but we adopt a more challenging setting. Specifically, we set $p_1 = 100$, $p_2 = 25$, $p_3 = 25$, $\rho_1 = 0.8$, $\rho_2 = 0.6$, $\alpha = 0.3$, and $\sigma_1 = \sigma_2 = 0.1$. We analyze the impact of sample sizes $n_1$ and $n_2$ on three methods by fixing $n_1 = 300$ and varying $n_2$ as $n_2 = k \times 100$ for $k = 1, \ldots, 4$. The experiments are repeated \num{30} times, and the mean RMSE per task is computed, with the results summarized in Figure \ref{fig:ablation}.
It is important to note that, due to the presence of distribution heterogeneity in this setting, HTL performs the worst. 

We analyze the ablation results from different perspectives. First, it is evident that both Ablation \num{3} and our proposed MTL-HMB outperform STL, Ablation \num{1}, and  Ablation \num{2}, indicating that Step \num{2} plays a crucial role in enhancing prediction performance. Second, by comparing Ablation \num{1} with STL, we observe that Ablation \num{1} consistently achieves lower loss across different sample sizes, demonstrating that Step \num{1} improves predictions for a single dataset. Third, when comparing Ablation \num{3} with our proposed method, Ablation \num{3} shows higher loss, suggesting that ignoring distribution heterogeneity in imputation negatively impacts performance. Fourth, we compare Ablation \num{1}, Ablation \num{2}, and our proposed MTL-HMB method, all of which incorporate Step \num{1}. The prediction results demonstrate that our method outperforms both Ablation \num{2} and Ablation \num{1}. This indicates that our MTL framework in Step \num{2} is more effective than hard parameter sharing, as it accounts for distribution heterogeneity, while hard parameter sharing performs better than STL. 
Fifth, even when comparing Ablation \num{2} with Ablation \num{3}—which uses a less effective imputation method—the latter still achieves better predictive performance. This further highlights the advantages of Step \num{2} over traditional MTL approaches.
Overall, the ablation experiments demonstrate that when both distribution and posterior heterogeneity are present, both steps of our proposed framework are crucial. 

\subsection{Comparison to Standard Block-wise Imputation Method}\label{A7}
In this section, we apply the statistical block-wise imputation method (MBI) proposed by \citet{xue2021integrating,xue2021statistical} to various simulation settings and the ADNI real dataset.
In particular, MBI does not account for distribution or posterior heterogeneity. It assumes that the relationships among all sources across different tasks are consistent, as well as the relationships between the sources and the response. The method first imputes all missing blocks and then constructs estimating equations based on the available information. These estimating equations are subsequently integrated to achieve efficient estimators.
We used the R package \texttt{BlockMissingData} to conduct the experiments, with the tuning parameters set to their default values. The RMSE was computed on a $20\%$ testing set.

In Sections \ref{sec: exp 2 tasks} and \ref{sec: exp multi}, we re-implemented the MBI method. MBI performed poorly across almost all settings, so we did not include its numerical results here. This poor performance can be attributed to several factors. First, MBI cannot handle nonlinear relationships and is limited to modeling linear relationships between sources and the response, which severely restricts its learning capacity. These findings underscore the substantial benefits of leveraging the encoder-decoder framework. Second, MBI is unable to address distribution or posterior heterogeneity.

To ensure a fair comparison, we reconsidered a linear data-generating process (DGP). Specifically, we modified the nonlinear DGP described in Section \ref{sec: exp 2 tasks} to a simpler linear DGP as follows:
\begin{align*}
y^{1} = \alpha \sum_{d=1}^{p} {v_{c,d}x^{1}_d}/{p} + (1-\alpha) \sum_{d=1}^{p} {v_{1,d}x^{1}_d}/{p} + \varepsilon_1,\\
y^{2} = \alpha \sum_{d=1}^{p} {v_{c,d}x^{2}_d}/{p} + (1-\alpha) \sum_{d=1}^{p} {v_{2,d}x^{2}_d}/{p} + \varepsilon_2,
\end{align*}
Where other parameters and settings remain unchanged, we evaluated the prediction performance of the four methods under this linear DGP. The results, presented in Table \ref{tab: linear}, show that MTL-HMB still achieves the best performance, followed by STL. HTL is constrained by distribution heterogeneity, while MBI, despite being designed for linear cases, suffers significant errors starting from the imputation step due to its assumption of no distribution or posterior heterogeneity. Consequently, its final predictions are notably poor.
\begin{table}[h]
\centering
\caption{Average RMSEs under linear setting.}
\begin{tabular}{cccc}
\toprule
 STL & HTL & MTL-HMB & MBI \\
\midrule
 $0.295 (0.028)$ & $0.765 (0.167)$ & $0.274 (0.029)$& $0.525 (0.296)$ \\
\bottomrule
\end{tabular}
\label{tab: linear}
\end{table}

Additionally, we evaluated MBI on the ADNI real dataset. The prediction results for Task 1 and Task 2 were $9.847 , (3.516)$ and $10.272 , (3.448)$, respectively. These findings further demonstrate the significant improvements brought by the encoder-decoder framework in real-world applications.
\subsection{Pseudo-code for Our Proposed MTL-HMB}\label{app: algorithm}
Algorithm \ref{alg} provides the pseudo-code for training our proposed MTL method. For simplicity, we set the mini-batch size to be the same across all $T$ datasets: $B^t = B$ for $t \in [T]$. 
For HBI, we divide the data into training and testing sets and train the parameters on the training set. Early stopping is applied to $\gL_{\text{pre}}$ on the $t$-th dataset's testing data to check for convergence and perform model selection.
For heterogeneous MTL, the data is split into training, validation, and testing sets. Parameters are trained on the training set, and the best hyperparameter combination is selected using the validation set. Early stopping is applied to $\gL_{\text{integ}}$ on the validation set to check for convergence, and the final prediction metrics are calculated on the test set.
In practice, we choose the regularization parameters $\gamma$, $\delta$, and $\kappa$ from the set $[0.01, 0.1, 1]$ for $\gR_{\text{orth}}$, $\gR_{\text{imp}}$, and $\gR_{\text{dr}}$. In our experiments, we found that the selection of $\gamma$, $\delta$, and $\kappa$ is robust, having minimal impact on the final prediction performance.
\newpage
{\spacingset{1} 
\begin{algorithm}[H]
\footnotesize
\begin{algorithmic}[1] 

\State Input: $T$ datasets denoted by $\{\vx_i^t,y_i^t\}_{i=1}^{n_t}$, where $\vx^t_i$ includes two blocks $\vx^t_{0,i}$ and $\vx^t_{t,i}$, learning rate $\eta$, mini-batch size for the $t$-th dataset is denoted by $B^t$.
\State \textbf{Step 1: HBI}
 \For{$t=1,\ldots,T$}\Comment{Imputation for task $t$-specific source}
 \State Initialize: $\mtheta^t$ (all parameters in this step)
 \While{not converged}
 \State Sample mini-batch of $B^t$ demonstrations from the $t$-th dataset $\{\vx_i^t,y_i^t\}_{i=1}^{n_t}$ and  mini-batch combination of $B^{-t}=\sum_{s\neq t} B^t$ demonstrations from the rest $T-1$ datasets.
 \For{$i=1,\ldots, B^t$} \Comment{Process batch from the $t$-th dataset.}
    \State $\vf^t_{i}=E_c(\vx^t_{0,i})$, $\vg^t_{i}=E_p^t(\vx^t_{0,i})$
    \EndFor
     \State Compute prediction loss $\gL_{\text{pre}}^t=\sum_{i=1}^{B^t}l(\vx^t_{t,i},G(\vf^t_{i}))$
    \State Compute reconstruction loss $\gL_{\text{recon}}^t=\sum_{i=1}^{B^{t}}l(\vx^{t}_{0,i},D(\vf^{t}_{i},\vg^{t}_{i}))$
    \For{$i=1,\ldots,B^{-t}$}\Comment{Process batch from the rest $T-1$ datasets.}
    \State $\vf^{-t}_{i}=E_c(\vx^{-t}_{0,i})$, $\vg^{-t}_{i}=E_p^{-t}(\vx^{-t}_{0,i})$
    \EndFor
    \State Compute reconstruction loss: $\gL_{\text{recon}}^{-t}=\sum_{i=1}^{B^{-t}}l(\vx^{-t}_{0,i},D(\vf^{-t}_{i},\vg^{-t}_{i}))$
    \State Parameter update $\mtheta^t\leftarrow\mtheta^t-\eta\nabla_{\mtheta^t}(\gL_{\text{pre}}^t+\gL_{\text{recon}}^t+\gL_{\text{recon}}^{-t})$
 \EndWhile
 \For{$i=1,\ldots,B^{-t}$}
    \State Imputation for task $t$-specific source: $\widehat\vx^{-t}_{t,i}=\widehat G(\widehat E_c(\vx^{-t}_{0,i}))$
    \EndFor
 \EndFor
\State Obtain samples with reconstructed features $\{(\vx^t_{0,i},\ldots, \widehat \vx^t_{t-1,i}, \vx^t_{t,i}, \widehat \vx^t_{t+1,i},\ldots,\widehat \vx^t_{T,i}), y^t_i)\}_{i=1}^{n_t}$
   \Statex
\State\textbf{Step 2: Heterogeneous MTL}
\State Initialize: $\mTheta$ (all parameters in this step)
\While{not converged}
\For{$t=1,\ldots,T$}
\For{$i=1\dots B^t$} \Comment{Process batch from the $r$-th dataset.}
    \State $\vh^{t}_{i} = \phi_c(\vx^t_{0,i})$, $\vk^{t}_{i} = \phi^{t}_p([\vx^t_{0,i}|\cdots| \widehat\vx^t_{t-1,i}|\vx^t_{t,i}|\widehat \vx^t_{t+1}|\cdots|\widehat \vx^t_{T,i}])$
    \State Set $\mH^t=[\vh^t_{i}\cdots\vh^t_{B^t}]^\top$, $\mK^t=[\vk^t_{i}\cdots\vk^t_{B^t}]^\top$
    \For{$l=1\dots L$}
    \If{$l==1$}
    \State $\bar\vh^t_{l,i}=\vh^{t}_{i}$, $\bar\vk^t_{l,i}=[\vh^{t}_{i}|\vk^{t}_{i}]$
    \Else 
    \State $\bar\vh^t_{l,i}=\vh^{t}_{l-1,i}$, $\bar\vk^t_{l,i}=[\vh^{t}_{l-1,i}|\vk^{t}_{l-1,i}]$
    \State $\vh^t_{l,i} = \text{Shared\_Path}(\bar\vh^t_{l,i})$,  $\vk^t_{l,i} = \text{Task\_Specific\_Path}^t(\bar\vk^t_{l,i})$
    \EndIf
    \EndFor
    \State  $\widehat{y}_i^{t} = g^t([\vh^t_{L,i}|\vk^t_{L,i}])$ 
    \EndFor
\EndFor
    \State Compute integration loss: $\gL_{\text{integ}} = \sum_{t=1}^T \sum_{i=1}^{B^t} l(y^t_i, \widehat y^t_i)$ 
    \State Compute orthogonal regularizer for features: $\gR_{\text{orth}} = \sum_{t=1}^T \|(\mH^{t})^{\top} \mK^{t}\|^2_F$
    \State Compute robust regularizer for imputation: $\gR_{\text{imp}} = \sum_{t=1}^T \sum_{s\neq 0, t}\|\mTheta^{t}_{s,p,1}\|^2_F$
    \State Compute regularizater for redundancy: $\gR_{\text{dr}} = \sum_{t=1}^T \sum_{l=1}^L \| (\mTheta^{t}_{c,l})^\top\mTheta^{t}_{p,l,1:d^{r}_{c,l-1}}\|_F^2$
    \Statex
    \State Parameters update: 
    \State $\mTheta \leftarrow \mTheta - \eta \nabla_{\mTheta} (\gL_{\text{integ}} + \gR_{\text{orth}} + \gR_{\text{imp}} + \gR_{\text{dr}})$
\EndWhile
\State \textbf{Output:} Learnt parameters $\mTheta$ 
\caption{Pseudo-code for Our Proposed MTL-HMB.} \label{alg}
\end{algorithmic} 
\end{algorithm}}
\subsection{Experimental details}\label{A5}
\subsubsection{Dataset description}\label{sec: real data des}
In this subsection, we provide a detailed description of the ADNI database used in Section \ref{sec: adni}. The ADNI study \citep{mueller2005alzheimer} aims to identify biomarkers that track the progression of Alzheimer’s disease (AD). The MMSE score, which measures cognitive impairment, is treated as the response variable, and we aim to select biomarkers from three complementary data sources: MRI, PET, and gene expression. 
Given the sparsity assumption, we use region of interest (ROI) level data rather than raw imaging data, as the latter might not be suitable for our method. MRI variables include volumes, cortical thickness, and surface areas, while PET features represent standard uptake value ratios (SUVR) of different ROIs. Gene expression variables are derived from blood samples and represent expression levels at different gene probes. To reduce the number of gene expression variables, we apply sure independence screening (SIS), narrowing it down to \num{300} variables. This results in a total of \num{680} features, including \num{267} MRI features and \num{113} PET features.
The data is sourced from ADNI-\num{2} at month \num{48}, where block-wise missingness occurs due to factors such as low-quality images or patient dropout. Using visit codes, we align MMSE with the imaging data to ensure they are measured within the same month.
Ultimately, we obtained two datasets: dataset \num{1} contains only MRI and PET sources, while dataset \num{2} includes MRI and gene expression sources. Both datasets have relatively small sample sizes, underscoring the importance of effectively using incomplete observations in the analysis.
\subsubsection{Implementation details and hyperparameter tuning}
In Section \ref{sec: exp}, we compare our proposed method (MTL-HMB) with Single Task Learning (STL) and Heterogeneous Transfer Learning (HTL). Here, we provide the implementation details of these three methods. For STL, we use standard deep neural networks to train each dataset individually. In contrast, HTL assumes no heterogeneity in the anchoring source and extracts task-shared representations from it, while task-specific representations are derived from task-specific sources. 

\textbf{STL.} For STL, we use standard deep neural networks to train each dataset individually. Each dataset is split into \num{60}\% for training, \num{20}\% for validation, and \num{20}\% for testing. On the training set, we perform hyperparameter tuning, including network width from \{\num{32}, \num{64}, \num{128}\}, depth from \{\num{2}, \num{3}, \num{4}, \num{5}\}, and batch size from \{\num{8}, \num{16}, \num{32}\} (with \num{8} included due to the smaller sample size in the ADNI database). Additionally, we set the learning rate to \num{0.001} and the early-stopping patience to \num{30}. 
To stabilize the optimization during iterations, we use the exponential scheduler \citep{patterson2017deep}, which decays the learning rate by a constant per epoch. In all numerical tasks, we set the decay constant to \num{0.95}, applied every \num{200} iterations. We tune the hyperparameters and select the best model on the validation set. Finally, the tuned hyperparameters are used to compute the prediction loss on the testing set.

\textbf{HTL.} For HTL, we adapt the network architecture from \citet{bica2022transfer} and modify it for our setting. Following their approach, the framework for handling heterogeneous feature spaces consists of a common encoder for shared source and task-specific encoders for task-specific sources, implemented using deep neural networks. The network widths are selected from \{\num{32}, \num{64}, \num{128}\} and depths from \{\num{2}, \num{3}, \num{4}\}. 
The remaining components are incorporated into an MTL network architecture, similar to the structure described in Section \ref{sec: integration}, where shared and task-specific pathways have depths chosen from \{\num{2}, \num{3}, \num{4}\}. The output dimensions of the first $L-1$ layers are selected from \{\num{32}, \num{64}, \num{128}\}, with the final layer predicting the corresponding response. The batch size is chosen from \{\num{8}, \num{16}, \num{32}\}, and the learning rate is set to \num{0.001}.
To remain consistent with \citet{bica2022transfer}, we train the prediction loss on the training set, along with regularization terms $\gR_{\text{orth}}$ and $\gR_{\text{dr}}$. Early stopping and hyperparameter tuning are performed based on the sum of the prediction losses across all datasets on the validation set, with an early-stopping patience of \num{30}. Finally, the tuned hyperparameters are used to compute the prediction loss on the testing set.

\textbf{MTL-HMB.} For the proposed MTL-HMB, we describe the method in two steps: Step 1 and Step 2.
\textbf{Step 1:} In HBI, the common encoder, task-specific encoders, decoder, and predictor use network architectures with widths selected from \{\num{8}, \num{16}, \num{32}\} and depths from \{\num{1}, \num{2}, \num{3}\}. The batch size is chosen from \{\num{8}, \num{16}, \num{32}\}, and the learning rate is set to \num{0.001}. Notably, since the features in our simulated data exhibit relatively simple linear relationships, we include smaller network widths and depths in our tuning.
\textbf{Step 2:}  To construct task-shared and task-specific mappings, the network architecture for the shared encoder $\phi_c$ and the task-specific encoders $\phi_p^t$ have widths selected from \{\num{32}, \num{64}, \num{128}\} and depths from \{\num{2}, \num{3}, \num{4}\}. The output dimensions are also chosen from \{\num{32}, \num{64}, \num{128}\}. For the prediction function, both shared and task-specific pathways have depths chosen from \{\num{2}, \num{3}, \num{4}\}, with the output dimensions of the first $L-1$ layers selected from \{\num{32}, \num{64}, \num{128}\}. The final layer predicts the corresponding response. The batch size is chosen from \{\num{8}, \num{16}, \num{32}\}, and the learning rate is set to \num{0.001}. Early stopping and hyperparameter tuning are conducted based on the sum of the prediction losses across all datasets on the validation set, using an early-stopping patience of \num{30}. Finally, the tuned hyperparameters are applied to compute the prediction loss on the testing set.
\subsection{Discussion about limitations and future work}\label{sec: future}
First, our proposed method essentially assumes that there is common information across all tasks that can be fused, which implies a relatively strong shared structure. For example, in Section \ref{sec: integration}, we assume the existence of a common mapping, $\psi_c$, between input features and responses for all $r \in [R]$. However, in reality, when there is strong heterogeneity across multiple datasets, the shared structure is often only partial. For instance, in three datasets, only two may share the common $\psi_c$, while the third task may be too heterogeneous to fuse with the first two. In such cases, an adaptive approach for MTL is needed, one that explores partially shared information among tasks while preserving the uniqueness of the highly heterogeneous task. Currently, some studies have considered adaptive MTL in relatively simple settings, such as linear cases \citep{duan2023adaptive,tian2023learning}. However, adaptive MTL in the presence of block-wise, distribution, and posterior heterogeneity remains unexplored, making it a meaningful direction for future research.

Moreover, it is worth noting that in both Section \ref{sec: imputation} (HBI) and Section \ref{sec: integration} (MTL), the hidden representations in each dataset are learned in two steps: the first step for imputation and the second step for learning the response. This process introduces some computational redundancy. A possible improvement would be to combine these two steps into one, unifying multiple tasks to learn the hidden representations for each task, which can then be used for both imputation and response learning. However, this approach poses computational challenges, such as how to balance different loss functions to achieve both accurate imputation and prediction. Thus, this remains a future research direction worth exploring.


\end{document}